\title{imaplus_report}
\author{Xinglin Xie}
\date{April 2025}

\documentclass[10pt,twocolumn,letterpaper]{article}

\usepackage{cvpr}
\usepackage{fancyhdr} 
\usepackage{amsmath, amsfonts, amssymb}
\usepackage{float}
\usepackage{adjustbox}
\usepackage{placeins}
\usepackage{graphicx} 
\usepackage{lipsum}
\usepackage{amsmath}

\usepackage{booktabs}
\usepackage{multirow}
\usepackage{makecell}
\usepackage{array}
\usepackage{tabularx}
\usepackage{ragged2e}
\usepackage{geometry}
\usepackage{indentfirst}
\usepackage[pagebackref,breaklinks,colorlinks]{hyperref}

\usepackage[capitalize]{cleveref}
\crefname{section}{Sec.}{Secs.}
\Crefname{section}{Section}{Sections}
\Crefname{table}{Table}{Tables}
\crefname{table}{Tab.}{Tabs.}

\begin{document}

\title{\textbf{STSeg - Complex Video Object Segmentation: The 1st Solution for 4th PVUW MOSE Challenge}}
\author{Kehuan Song$^{1}$\hspace{1em}  Xinglin Xie$^{1}$\hspace{1em}  Kexin Zhang$^{1}$\hspace{1em}  Licheng Jiao$^{1}$\hspace{1em}  Lingling Li$^{1}$\hspace{1em}  Shuyuan Yang$^{1}$\\ \\
        $^{1}$Xidian University
        } 



\maketitle
\begin{abstract}
    Segmentation of video objects in complex scenarios is highly challenging, and the MOSE\cite{ding2023mose} dataset has significantly contributed to the development of this field. This technical report details the STSeg solution proposed by the "imaplus" team.By finetuning SAM2\cite{ravi2024sam} and the unsupervised model TMO\cite{cho2023treating} on the MOSE dataset, the STSeg solution demonstrates remarkable advantages in handling complex object motions and long-video sequences. In the inference phase, an Adaptive Pseudo-labels Guided Model Refinement Pipeline is adopted to intelligently select appropriate models for processing each video. Through finetuning the models and employing the Adaptive Pseudo-labels Guided Model Refinement Pipeline in the inference phase, achieving a J\&F score of 87.26\% on the test set of the 2025 4th PVUW Challenge MOSE Track, securing the 1st place and advancing the technology for video object segmentation in complex scenarios.
\end{abstract}

\section*{1. Introduction}
The PVUW Challenge has organized a competition centered around complex video segmentation\cite{Ding2024PVUW}. This competition is divided into two tracks: one is the Complex Video Object Segmentation Track based on the MOSE dataset, and the other is the Motion Expression guided Video Segmentation track based on the MeViS\cite{Ding_2023_ICCV} dataset. In this paper, we share our approach for the first track.Video object segmentation (VOS), an extension of image object segmentation, aims to segment specific objects from a sequence of video frames. This competition requires the first frame mask of the target object as input to complete the segmentation of subsequent frames. This task is highly challenging due to the diverse and complex appearance changes of objects in videos\cite{cheng2023putting}\cite{cheng2021modular}, accurately address significant changes in object appearance in subsequent frames. Furthermore, accurately inferring the occluded part's mask based on limited visible information and maintaining segmentation consistency before and after occlusion is a key issue that needs to be resolved urgently.

Current mainstream VOS methods often embed predictions from past frames into memory \cite{seong2020kernelized}\cite{oh2019video}\cite{zhang2023joint}\cite{cheng2021rethinking}and use attention mechanisms to associate memory frame representations with current frame features, thereby expanding contextual information. For example, XMem\cite{cheng2022xmem} achieves a J\&F score of 87.7\% on DAVIS2017\cite{pont20172017} and 86.1\% on YouTube-VOS\cite{xu2018youtube}. These methods effectively store and utilize features from past frames in memory modules, allowing models to consider multi-frame information over time when processing current frames, thereby making more accurate judgments about object appearance changes and motion trajectories. However, such pixel-level matching methods are prone to noise interference in complex environments, especially in the presence of occlusion and other factors, leading to significant performance degradation. The re-emergence of objects and the presence of confusing objects (similar neighboring objects) are the main causes of tracking failures. 

To promote the application of VOS in real-world scenarios, the competition has constructed the MOSE dataset for complex video object segmentation, which is used to study object tracking and segmentation in complex environments. The MOSE dataset features complex scenes with severe occlusion, disappearance, and re-emergence of objects, as well as small-scale issues\cite{videnovic2024distractor}, which not only challenge object segmentation in images but also greatly increase the difficulty of tracking occluded objects over time. Methods that perform well on mainstream VOS datasets show significant performance drops on the MOSE dataset. For instance, XMem without finetuning achieves a J\&F metric of 87.7\% on DAVIS 2017 but only 57.6\% on MOSE.

To effectively address these challenges, models need to possess strong image representation and long-term memory storage capabilities. In the VOS domain, SAM2 and the unsupervised model TMO each have their own strengths. SAM2 has revealed the negative impact of visual redundancy in large-capacity memory on localization capabilities and the positive role of temporal tagging in improving tracking performance. TMO, through its motion-as-option network and collaborative network learning strategy, can reduce reliance on motion cues and improve segmentation performance in complex scenarios.

Based on these insights, this report proposes the STSeg solution, which integrates the advantageous modules of SAM2 and TMO. We finetuned the SAM2,  and TMO models on the MOSE dataset and then introduced the Adaptive Pseudo-labels Guided Model Refinement Pipeline module. The system is able to dynamically assign tasks to the most suitable model for each video, thereby enhancing the efficiency and accuracy of video frame segmentation and tracking tasks. STSeg achieved 1st place in the MOSE Track of the 4th PVUW challenge 2025, demonstrating its superior performance and robustness in complex scenarios.
\section*{2. STSeg Solution}

We optimized the proposed solution in the training finetuning and inference stages. In the training stage, we finetuned the SAM2 and TMO models on the MOSE dataset to make them more suitable for the requirements of video object segmentation tasks in complex scenarios.In the inference stage, we employed five models, namely SAM2, TMO, cutie, XMem, and LiVOS, to conduct inference operations on the MOSE test set. We constructed comprehensive pseudo - labels using the mask annotations generated by the inference of each model. Based on the generated pseudo - labels, we intelligently selected the model with the optimal performance for different video contents. 

\subsection*{2.1 Model finetuning}
\subsubsection*{2.1.1 Finetuning SAM2}
SAM2, relying on the memory attention module, can correct predictions based on the context of object memories from previous frames, enabling stable segmentation of objects in videos. Meanwhile, it possesses a powerful zero - shot generalization ability, allowing it to operate on unseen data. However, the MOSE dataset presents numerous complex scenarios such as severe occlusions, disappearances and reappearances, small targets, and confusions among similar objects. Simply using SAM2 for inference cannot achieve satisfactory results. Therefore, we finetune SAM2 using the MOSE dataset.

\begin{figure}
    \centering
    \begin{subfigure}[t]{0.9\linewidth}
        \centering
        \includegraphics[width = 0.3\linewidth]{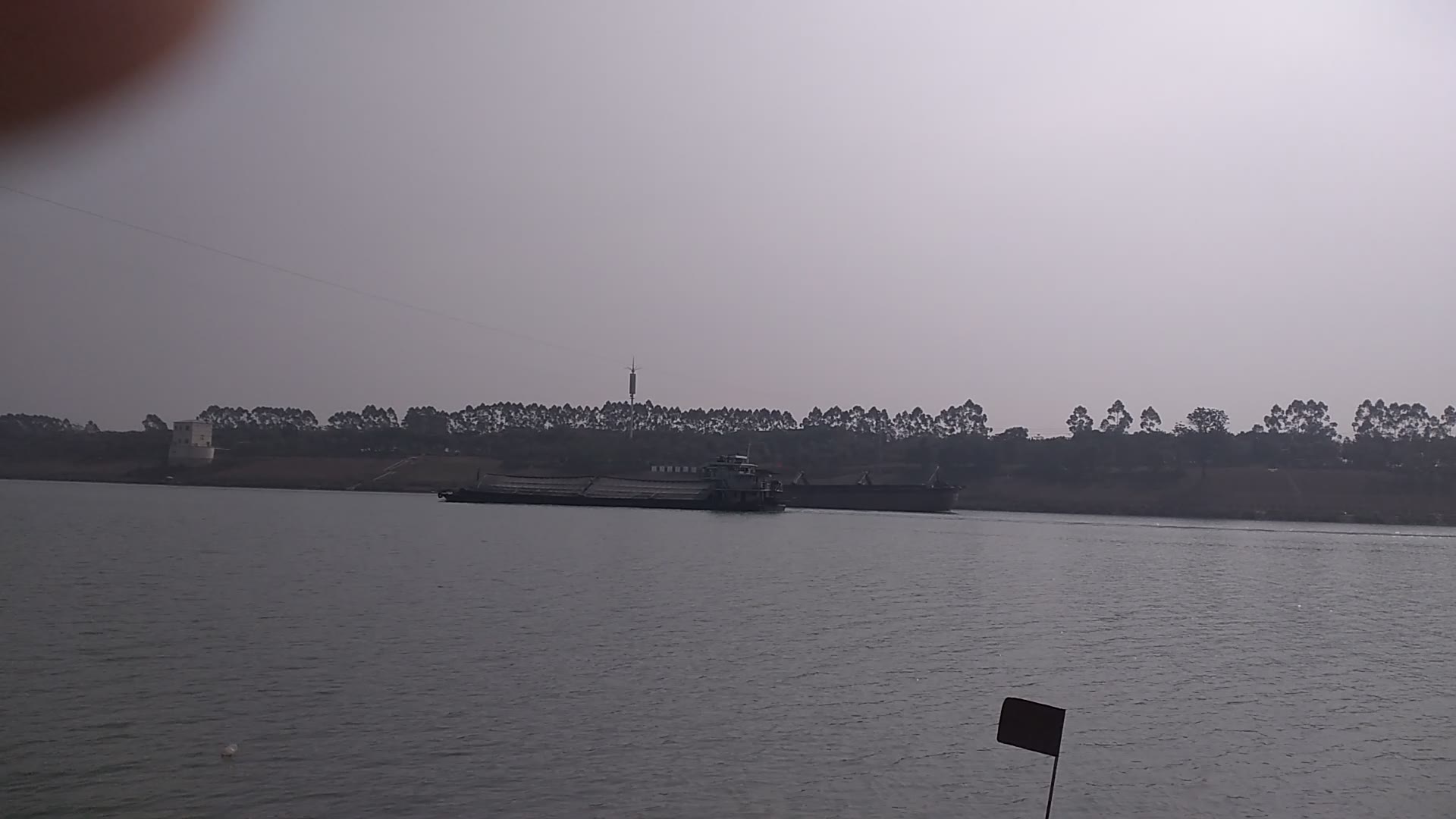}
        \includegraphics[width = 0.3\linewidth]{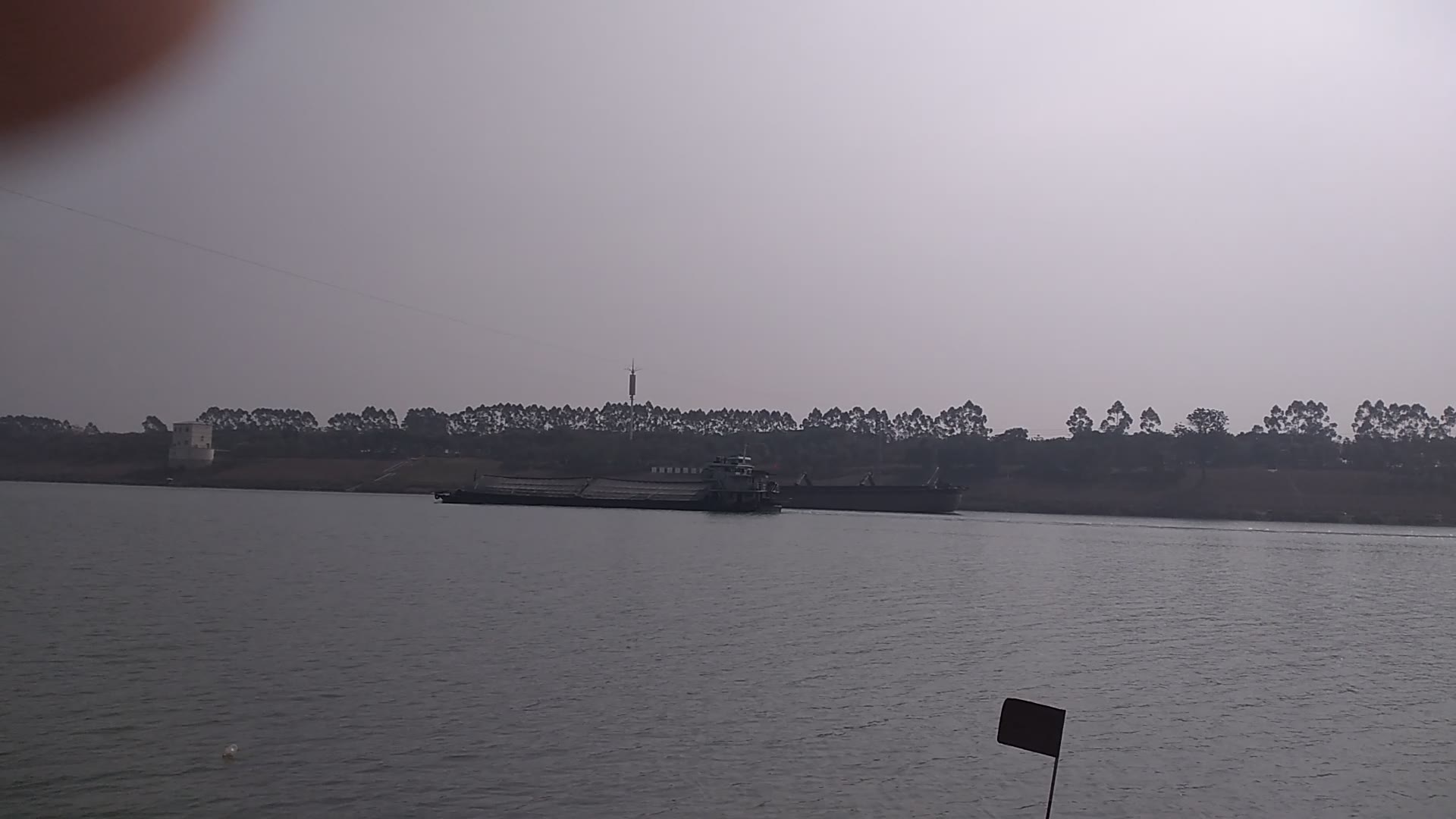}
        \includegraphics[width = 0.3\linewidth]{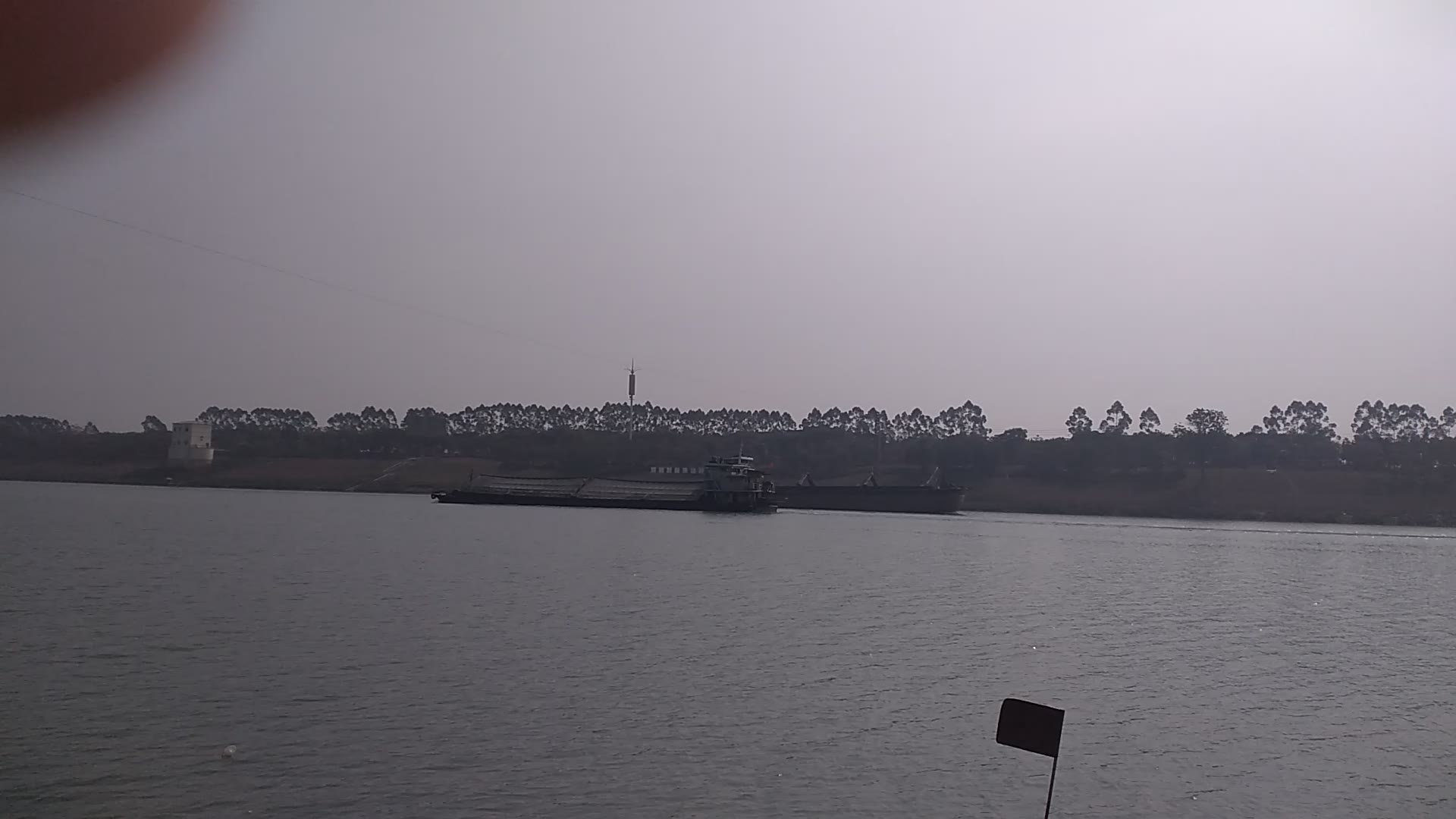}
        \caption{}
        \label{fig:sam-a}
    \end{subfigure}

    \begin{subfigure}[t]{0.9\linewidth}
        \centering
        \includegraphics[width = 0.3\linewidth]{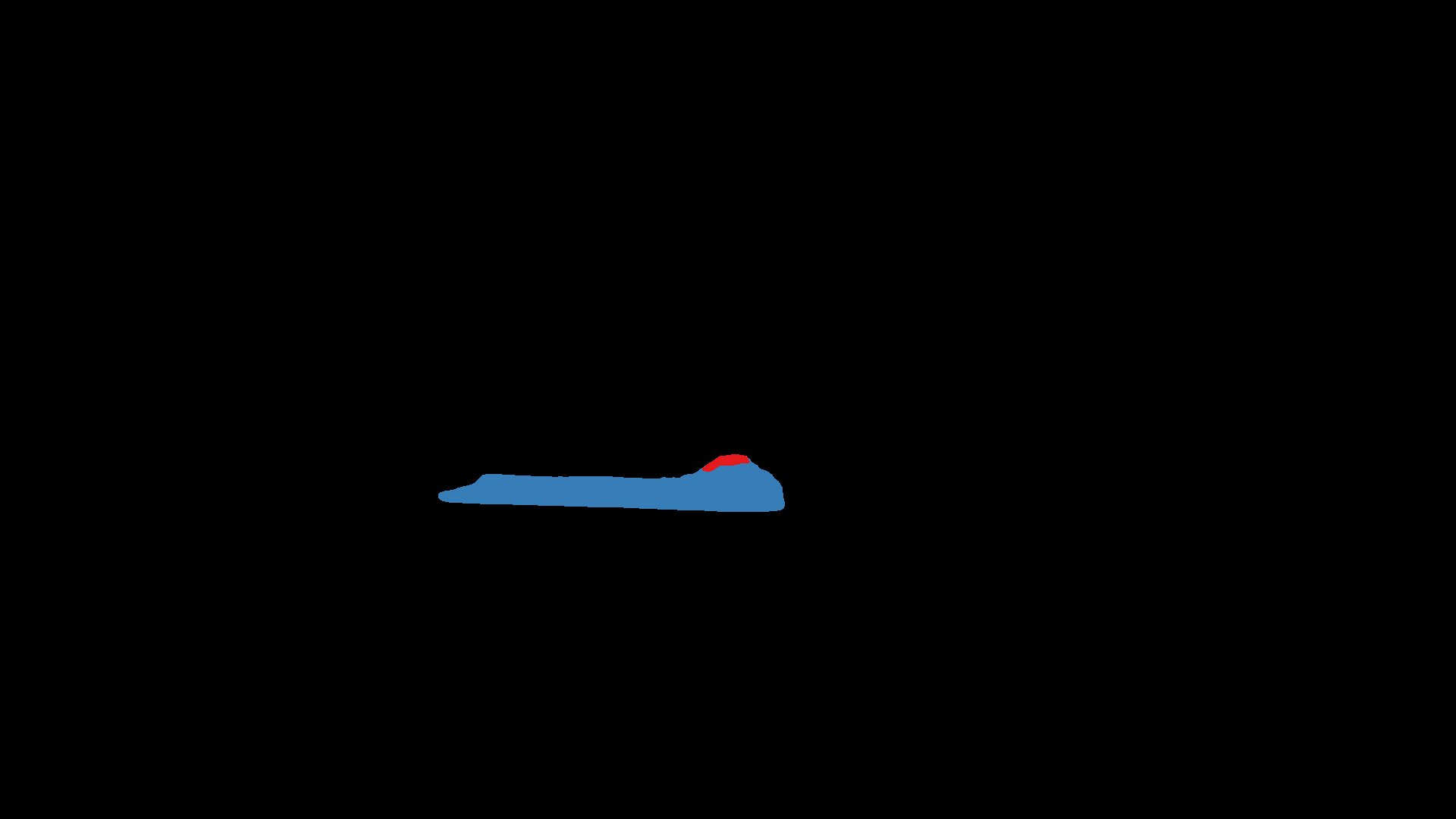}
        \includegraphics[width = 0.3\linewidth]{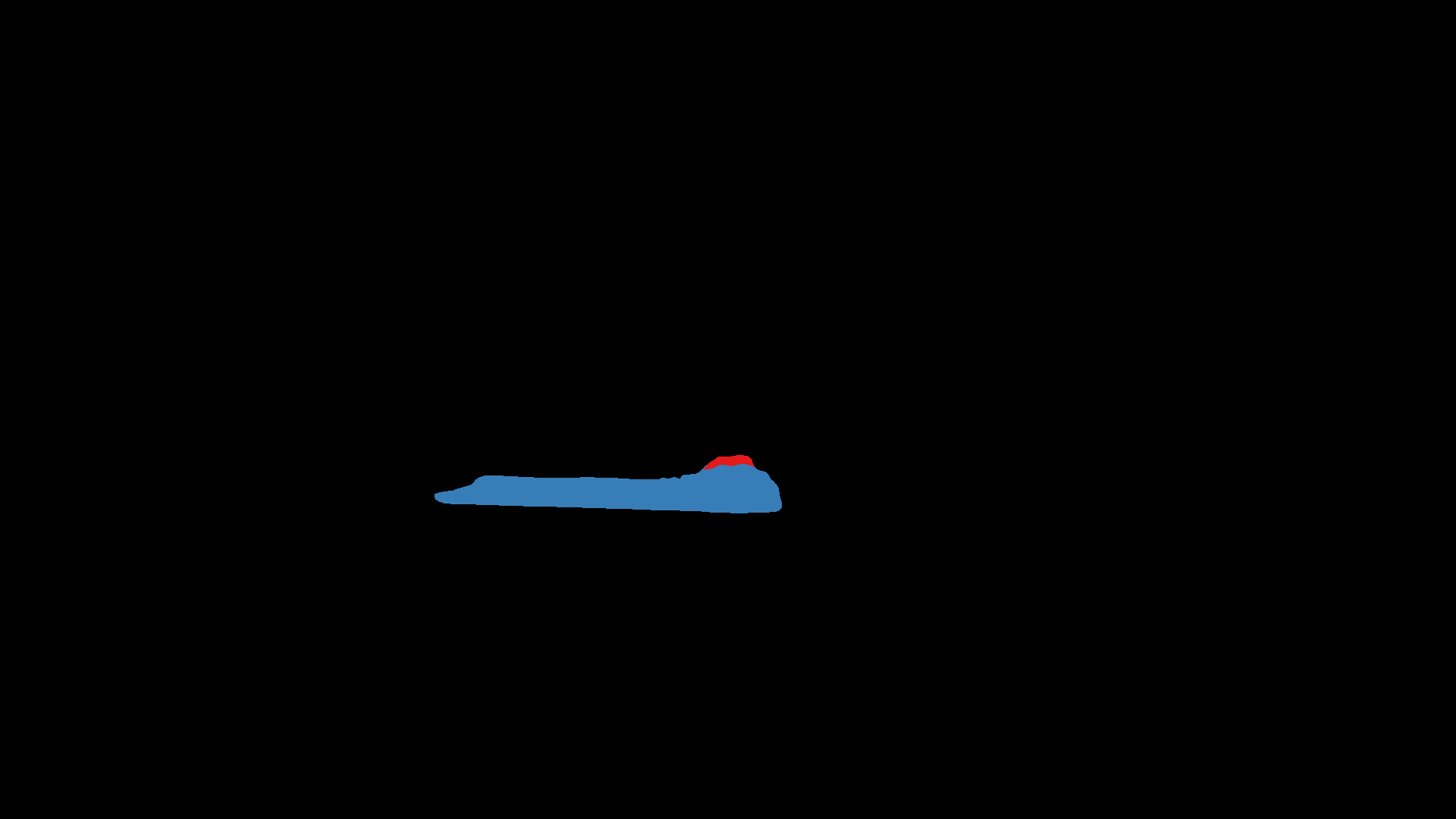}
        \includegraphics[width = 0.3\linewidth]{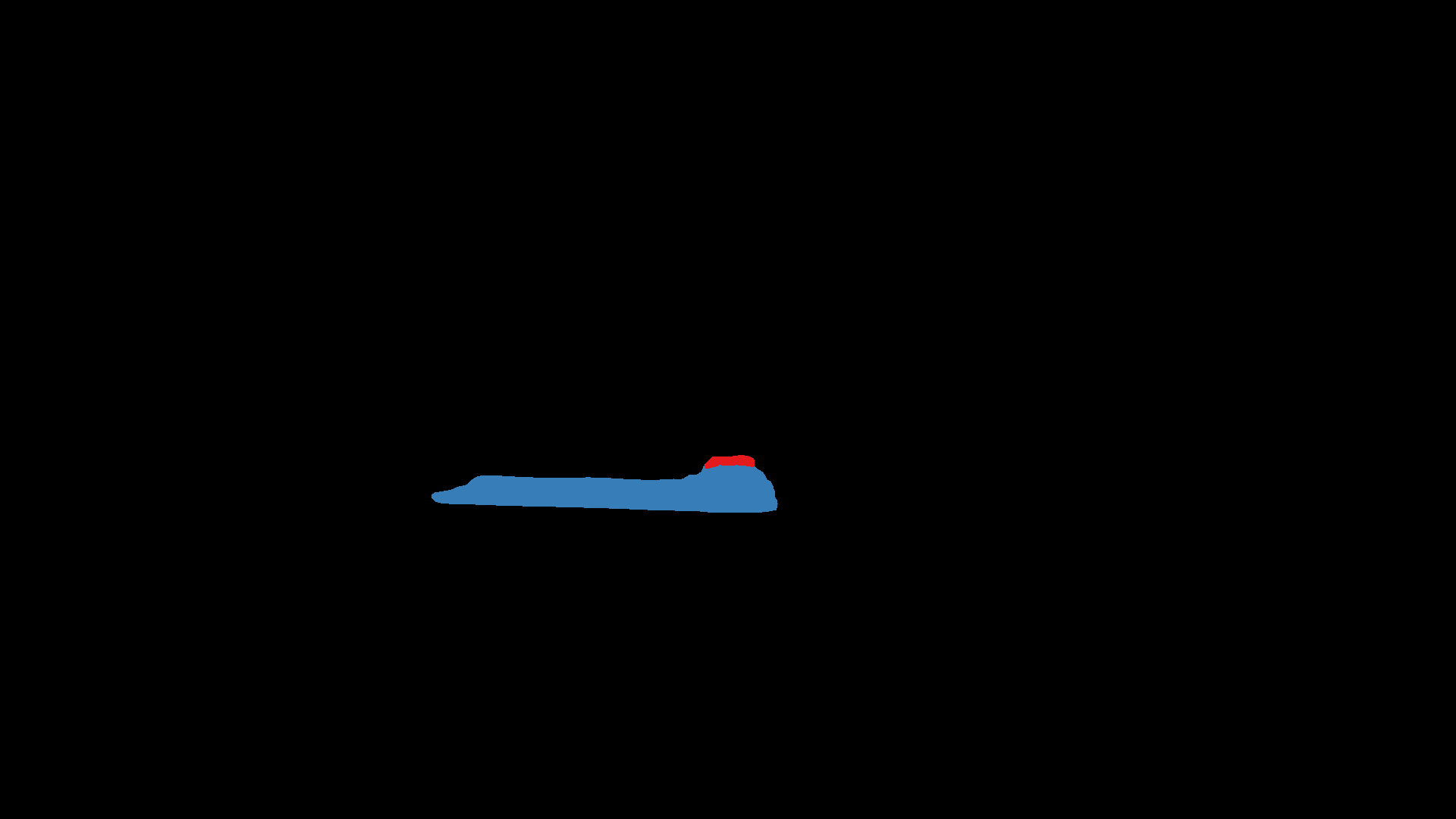}
        \caption{}
        \label{fig:sam-b}
    \end{subfigure}

    \begin{subfigure}[t]{0.9\linewidth}
        \centering
        \includegraphics[width = 0.3\linewidth]{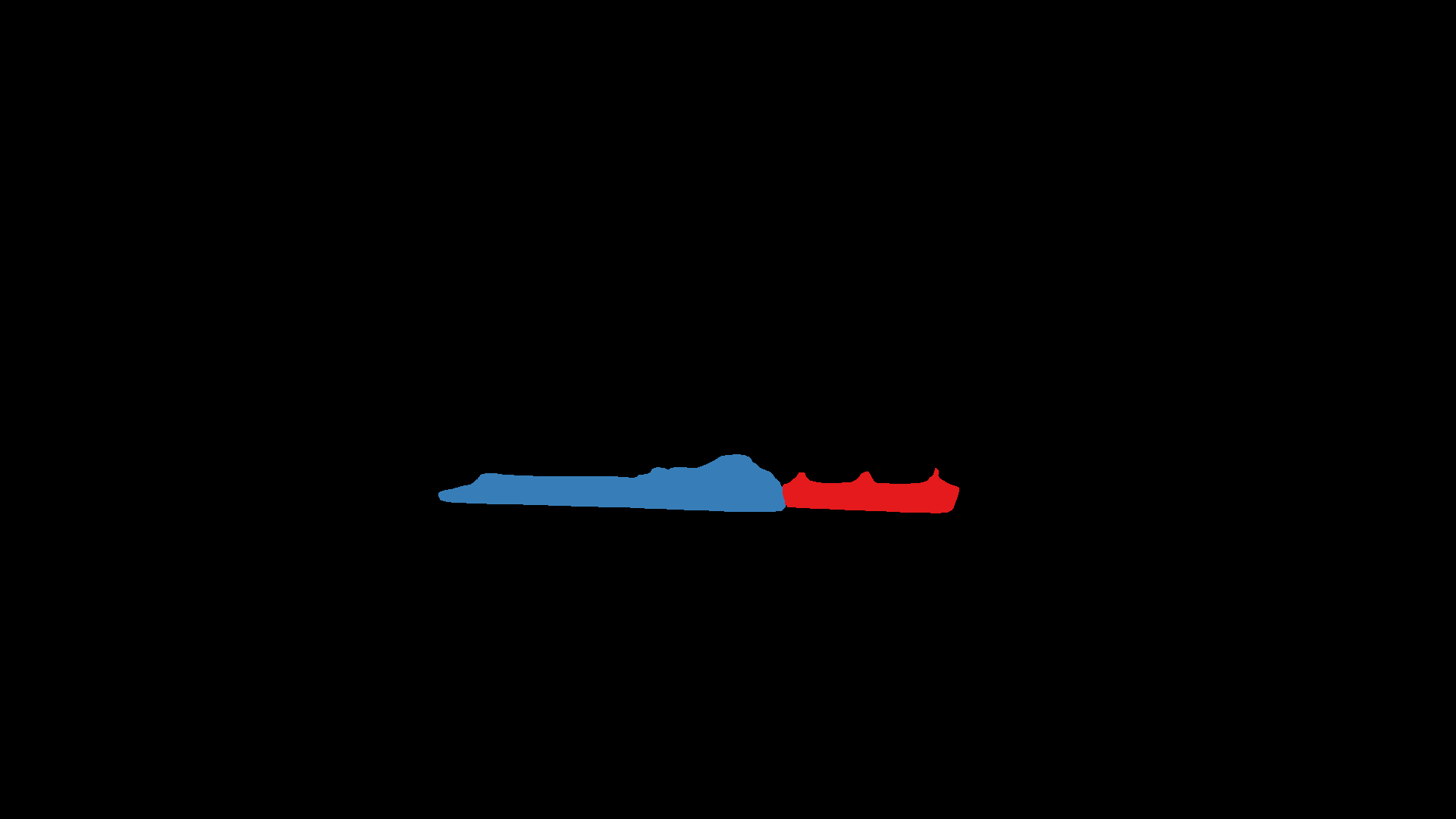}
        \includegraphics[width = 0.3\linewidth]{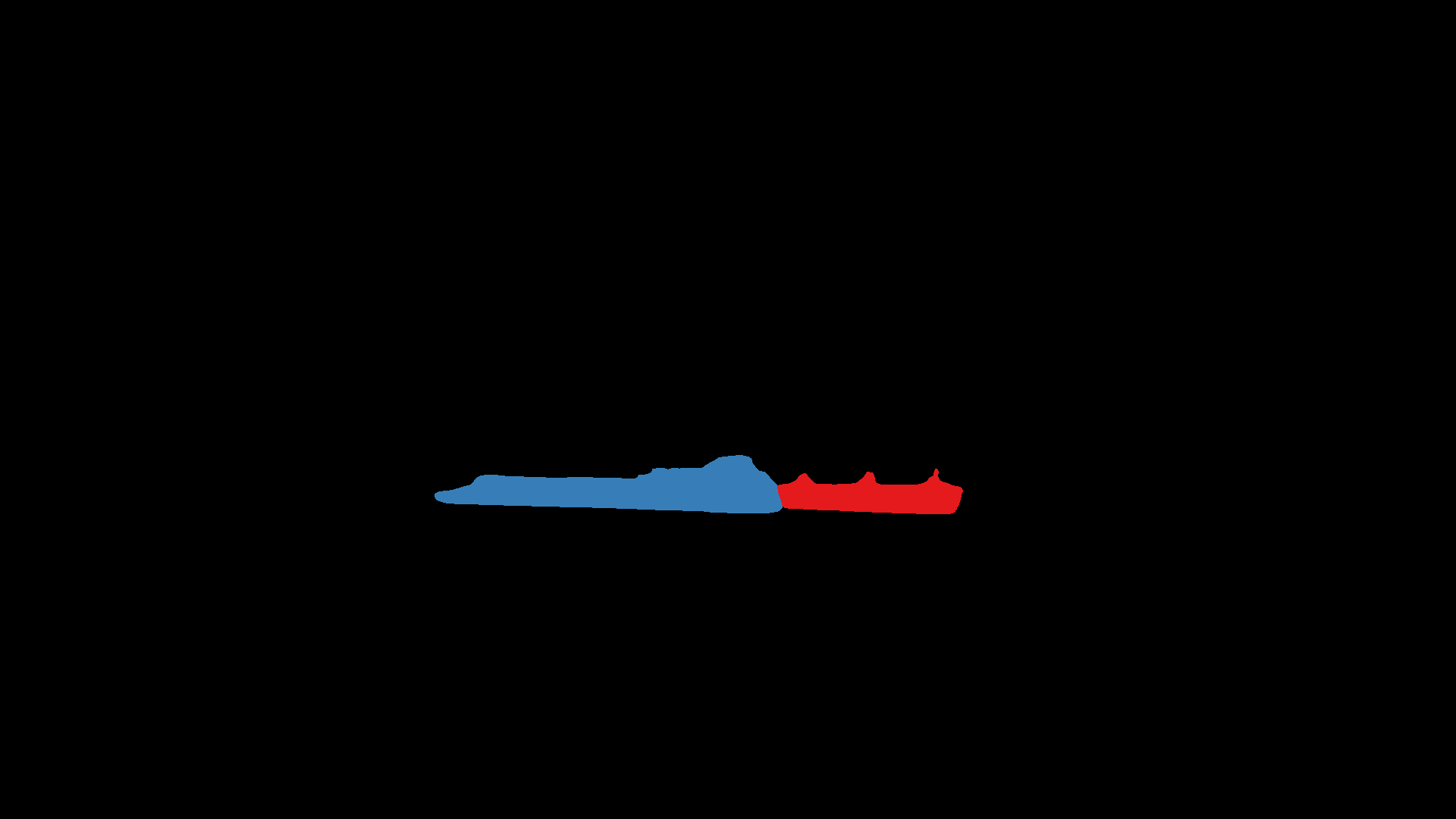}
        \includegraphics[width = 0.3\linewidth]{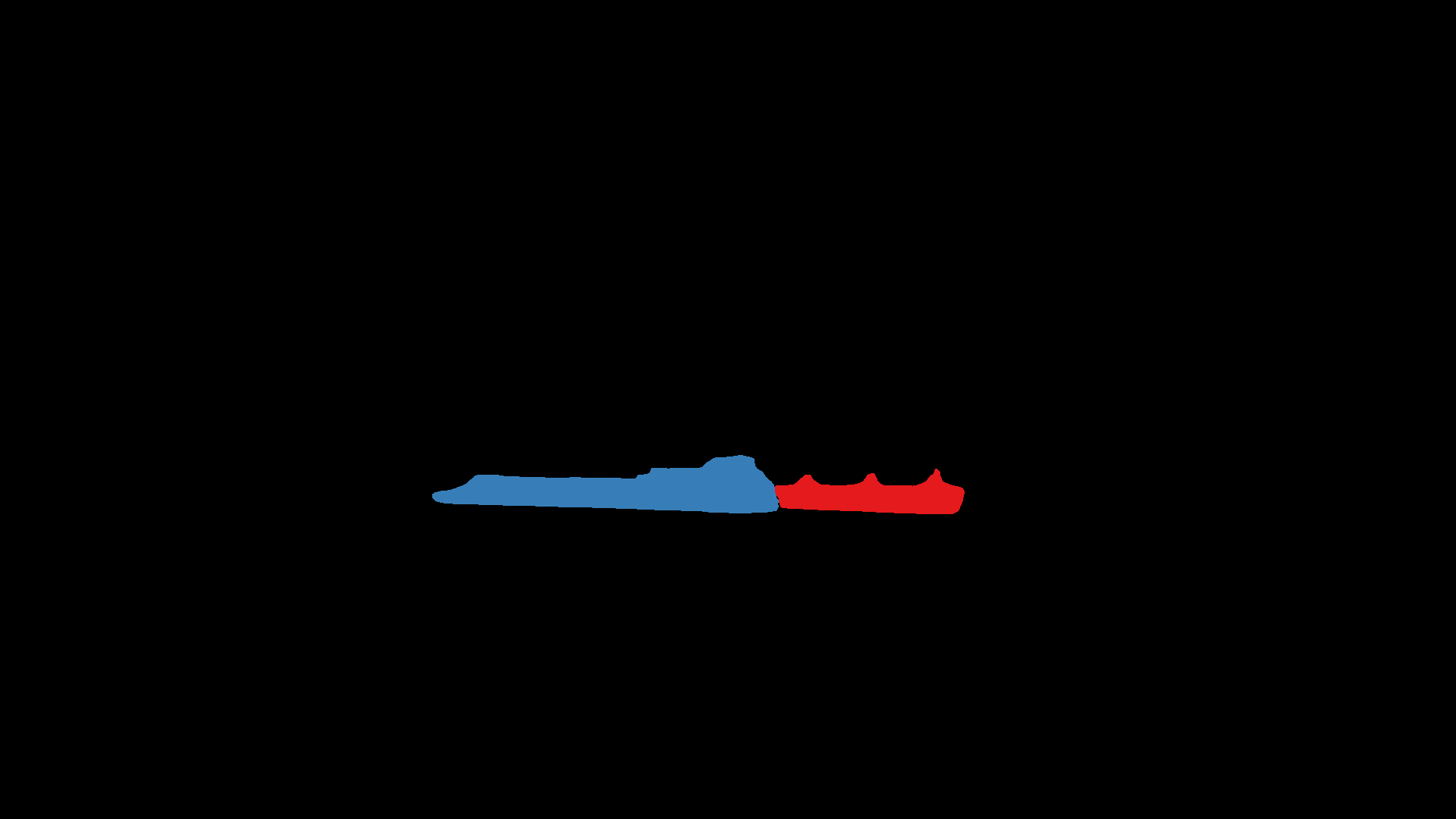}
        \caption{}
        \label{fig:sam-c}
    \end{subfigure}

    \caption{(a) shows the original images of the MOSE test set.
(b) presents that SAM2 without finetuning (large2.1 checkpoint) fails to track it successfully. (c) displays that SAM2 with fine-tuning can effectively track the occluded dragon boat when it reappears.}
    \label{fig:sam} 
\end{figure}

To enhance the performance on the MOSE dataset while maintaining the strong generalization ability of SAM2, we freeze the encoder part to retain its powerful feature extraction capability and conduct targeted finetuning on the decoder. We take 85\% of the MOSE training set as the training data and feed it into the model. To reduce the training time, we use the large checkpoints of checkpoint 2.1 as the pre-trained weights and train for 40 epochs using 8 A100 GPUs each having 80 GB of memory. We configure the hyperparameters, setting the base\_lr to$5\times10^{- 6}$, the vision\_lr to $3\times10^{- 6}$, and define the optimizer as Adam. During the training process, we iteratively process the training data, calculate the Focal loss, Dice loss, and IoU loss, and obtain the total loss through weighted summation. We update the model weights through gradient accumulation and backpropagation, and save the model every 5 epochs. We use 15\% of the MOSE training set as the validation set and test the J\&F score of each saved model on the validation set. We present some quantitative results before and after finetuning in Fig\ref{fig:sam}
\begin{figure*}[htbp]
    \centering
    \includegraphics[width=0.9\textwidth]{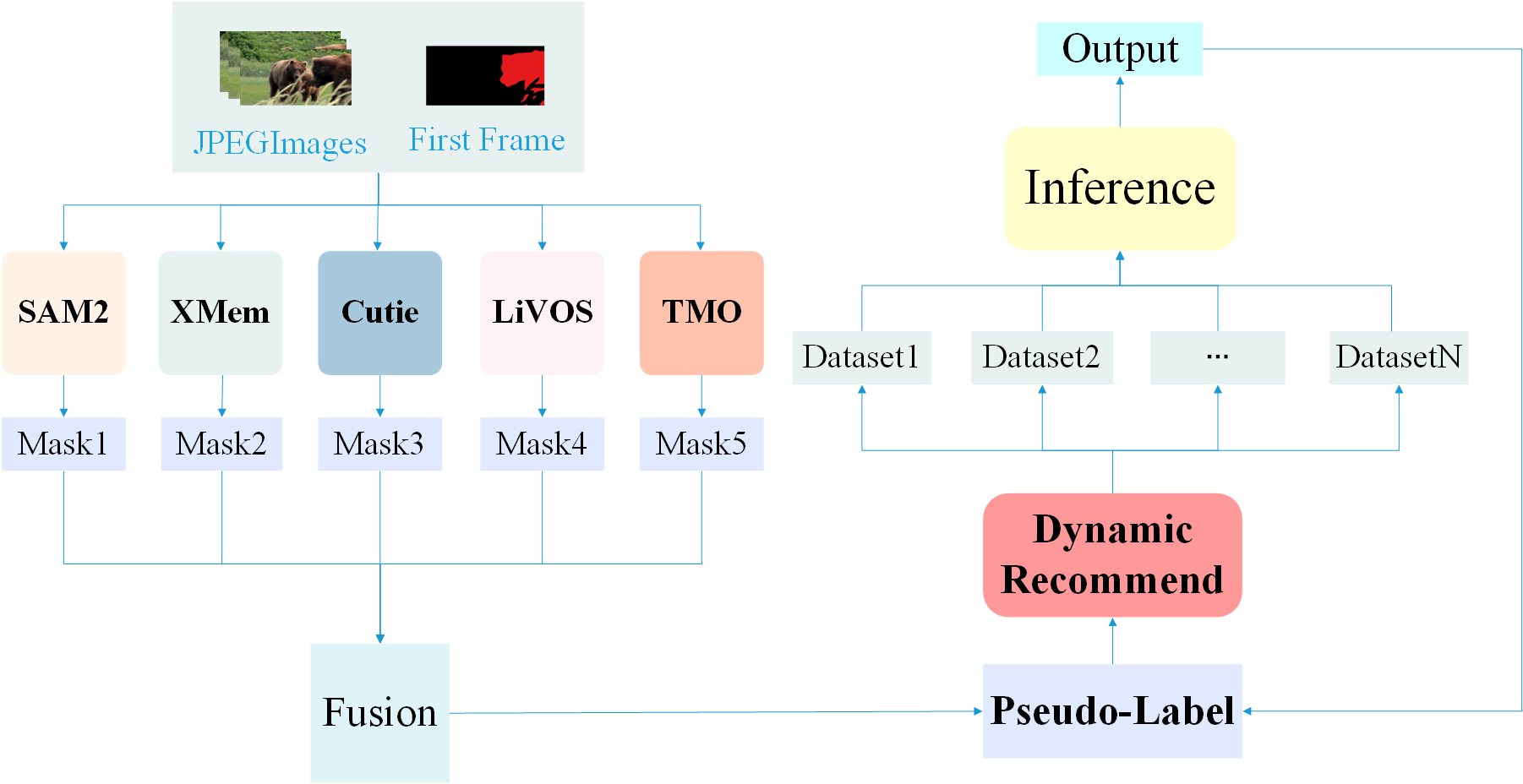} 
    \caption{Overview of the PGMR Framework. Inference and Pseudo-Label-Based Model Selection on MOSE Test Set: Employing five models (SAM2, TMO, Cutie, XMem, and LiVOS) to conduct inference operations. Comprehensive pseudo-labels are constructed using the mask annotations from each model's inference. Based on these pseudo-labels, the model with optimal performance for different video contents is intelligently selected.} 
    \label{fig:PGMR_framework} 
\end{figure*}

\subsubsection*{2.1.2 Finetuning TMO}
TMO is a motion-a-option network used for unsupervised video object segmentation (VOS), combined with a collaborative network learning strategy. Its aim is to reduce the dependence of existing methods on motion signals and to improve segmentation performance in complex scenarios. The motion-as-option network extracts appearance and motion features separately through independent encoders. Specifically, appearance features are obtained by processing RGB images, and motion features can be acquired from optical flow maps. Then, the features are fused to generate the final features, thereby reducing the dependence on motion cues.

When finetuning the unsupervised model on the MOSE dataset, we apply data augmentation techniques to the training set, including enlarging the images to highlight small targets. During the fine-tuning process, we first use the RAFT model to generate optical flow maps for the MOSE dataset. Then, we uniformly adjusted the size of the images and optical flow maps to meet the input requirements of the model. At the same time, we use nearest-neighbor interpolation for the segmentation masks to maintain the values of either 0 or 1. We adjust the model parameters, use the cross-entropy loss function and the Adam optimizer, set the learning rate to $1\times10^{- 5}$, and the batch size to 16, and freeze all batch normalization layers.

\subsection*{2.2 Adaptive Pseudo-labels Guided Model Refinement Pipeline}
After analyzing the dataset, we found it challenging to achieve good results in all scenarios using a single model. Therefore, we propose an Adaptive Pseudo-labels Guided Model Refinement Pipeline (PGMR), as shown in Fig \ref{fig:PGMR_framework}with specific implementation steps as follows:

\subsubsection*{2.2.1 Multi-Model Inference: Independent Processing and Result Collection}
In video frame segmentation and tracking tasks, we first employ multi-model independent inference to process the same set of video frames. Each model demonstrates unique performance advantages in different scenarios based on its design features and training data. To fully leverage the strengths of each model, we have designed a parallel inference framework that ensures each model can operate independently and produce results without interference from other models. This framework allows multiple models to perform inferences on the same set of video frames simultaneously, enabling each model to perform at its best without being influenced by others.
The output results of each model are collected separately and include segmentation masks, tracking IDs, and confidence scores. Segmentation masks are used to accurately delineate the boundaries of target objects within video frames while tracking IDs are employed to continuously track the positional changes of target objects throughout the video sequence and confidence scores reflect the model's assessment of each prediction. 

\subsubsection*{2.2.2 Pseudo-Label Fusion: Generating a Baseline Result}
To optimize the performance of video frame segmentation and tracking tasks, it is crucial to integrate the inference results of multiple models into a comprehensive pseudo-label. This pseudo-label serves as a key baseline for the subsequent optimization process and helps identify the model that performs optimally for different video contents. The generation of the pseudo-label involves several detailed steps:

\begin{itemize}
    \item Firstly, a consistency check is carried out by comparing the segmentation masks and tracking IDs of different models to identify the regions where the model results are consistent and those where they are inconsistent.
    \item Subsequently, confidence weighting is performed. Weights are assigned to each model based on its historical performance and the confidence scores associated with its predictions. 
    \item Finally, a voting mechanism is employed for the regions where the models produce conflicting results, and a conflict resolution strategy is adopted. 
\end{itemize}

\begin{figure}[H]
    \centering
    \begin{subfigure}[b]{0.3\linewidth}
        \centering
        \includegraphics[width=\linewidth]{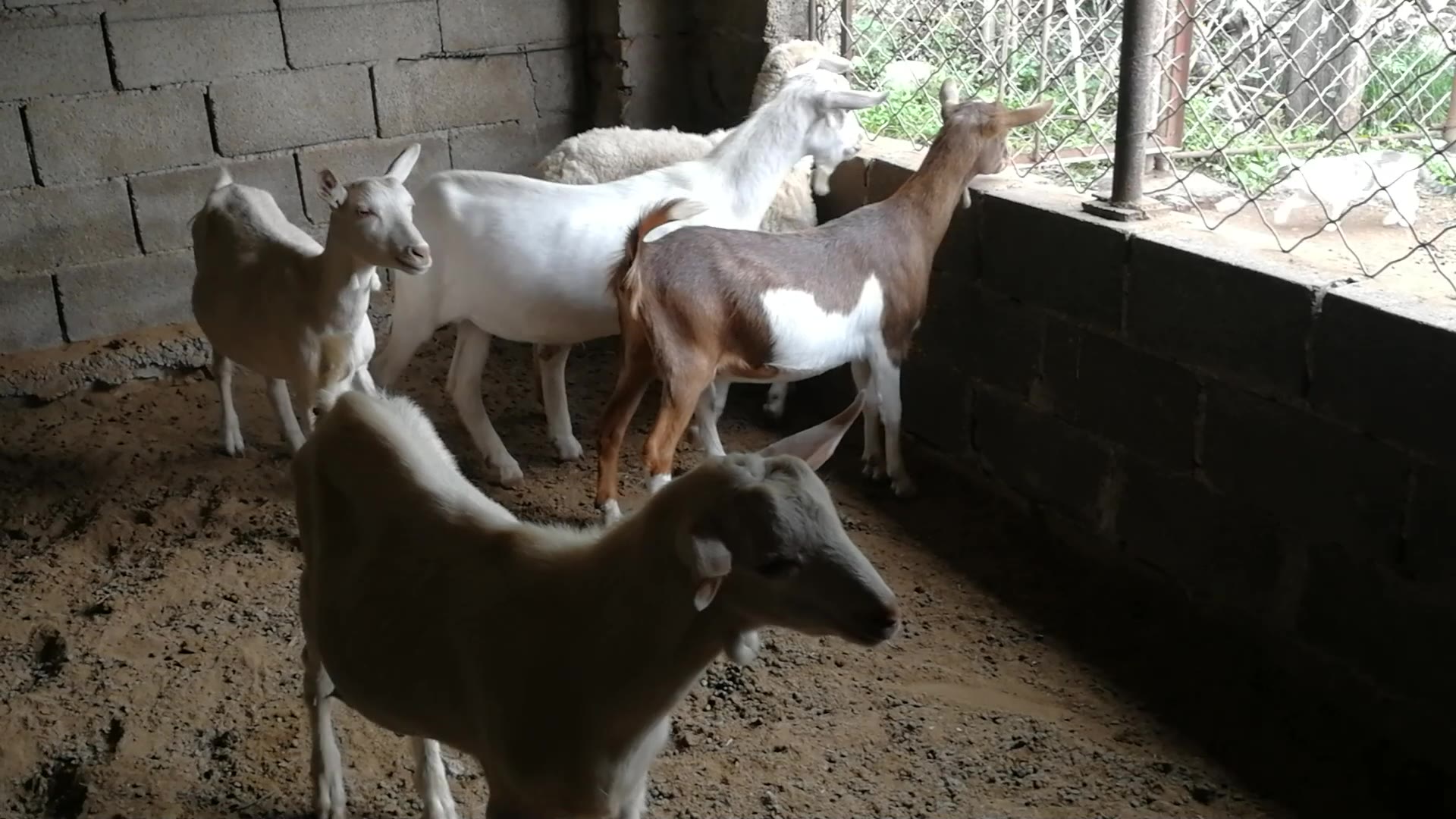}
        \caption{}
    \end{subfigure}
    \hfill
    \begin{subfigure}[b]{0.3\linewidth}
        \centering
        \includegraphics[width=\linewidth]{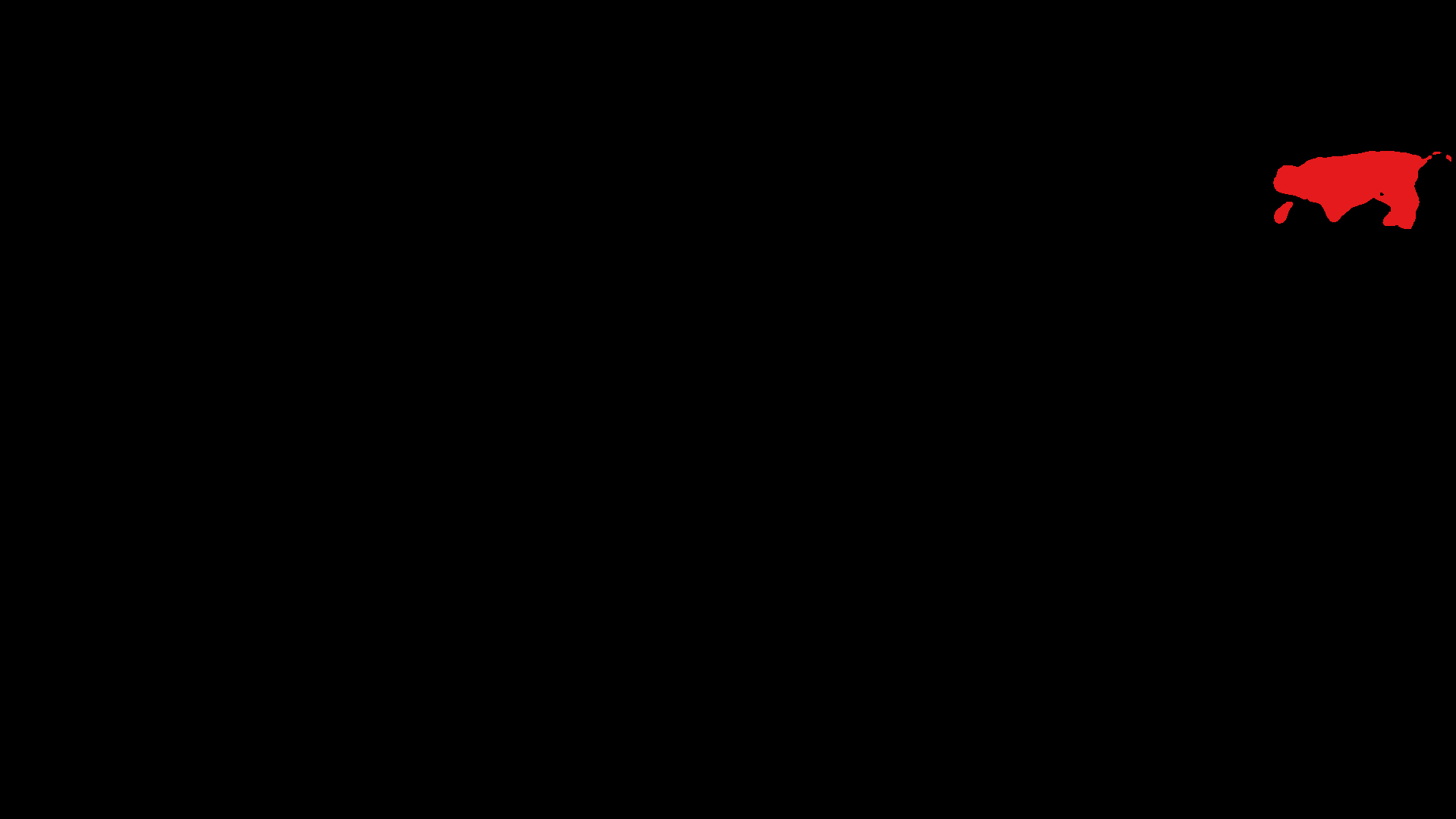}
        \caption{}
        \label{fig:sub-b}
    \end{subfigure}
    \hfill
    \begin{subfigure}[b]{0.3\linewidth}
        \centering
        \includegraphics[width=\linewidth]{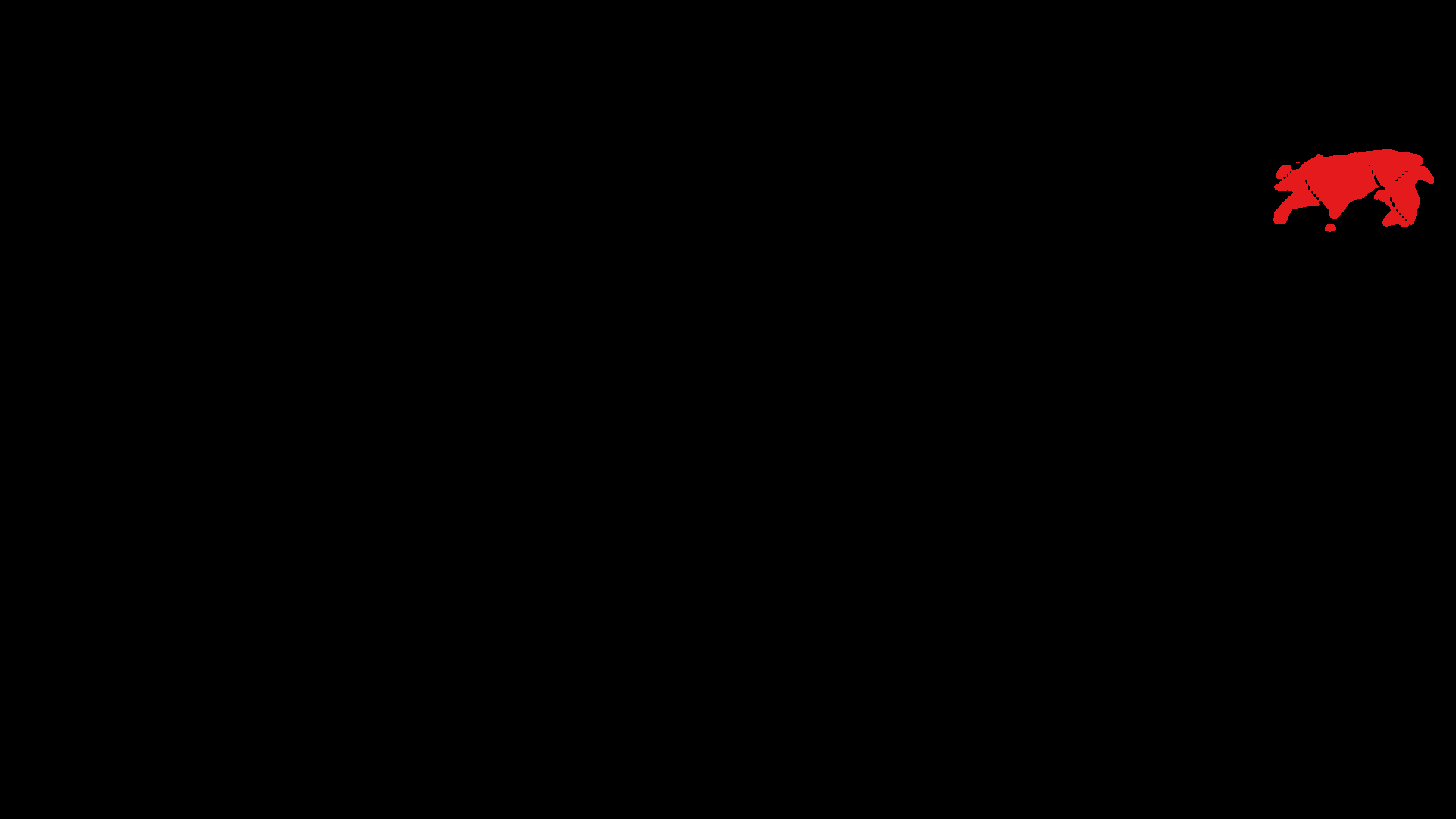}
        \caption{}
        \label{fig:sub-c}
    \end{subfigure}
    \vspace{0.1cm} 

    \begin{subfigure}[b]{0.3\linewidth}
        \centering
        \includegraphics[width=\linewidth]{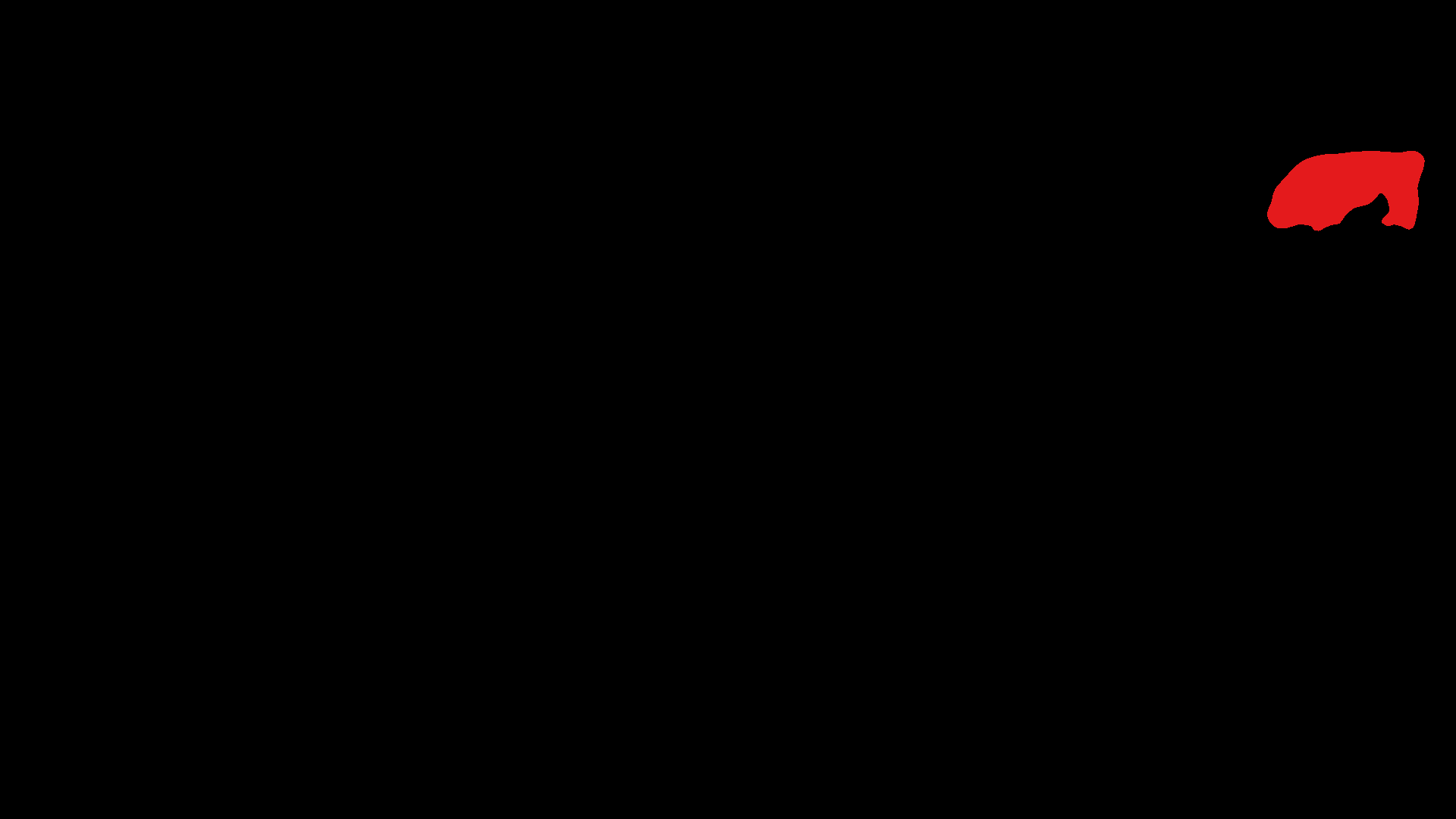}
        \caption{}
        \label{fig:sub-d}
    \end{subfigure}
    \hfill
    \begin{subfigure}[b]{0.3\linewidth}
        \centering
        \includegraphics[width=\linewidth]{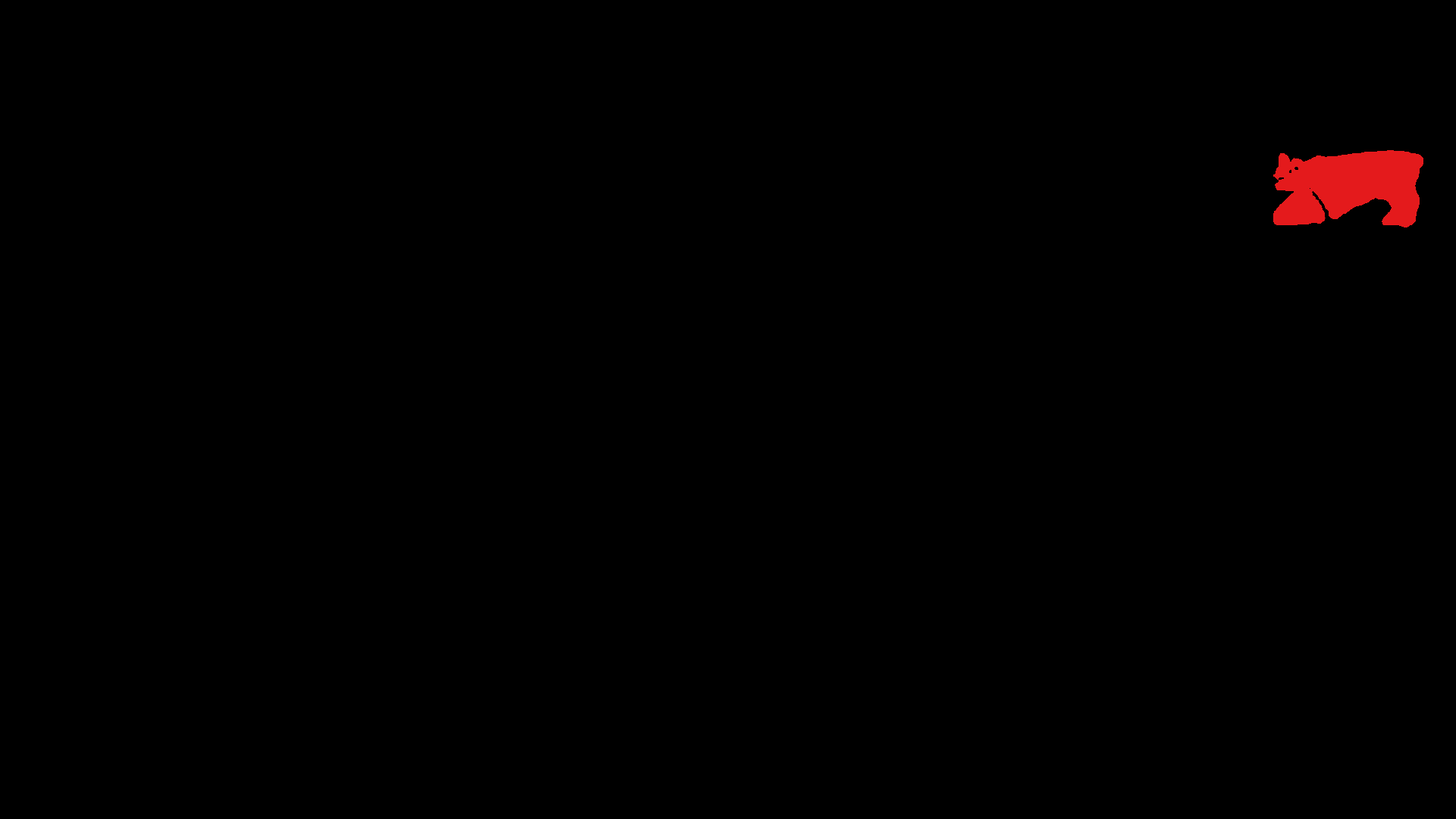}
        \caption{}
        \label{fig:sub-e}
    \end{subfigure}
    \hfill
    \begin{subfigure}[b]{0.3\linewidth}
        \centering
        \includegraphics[width=\linewidth]{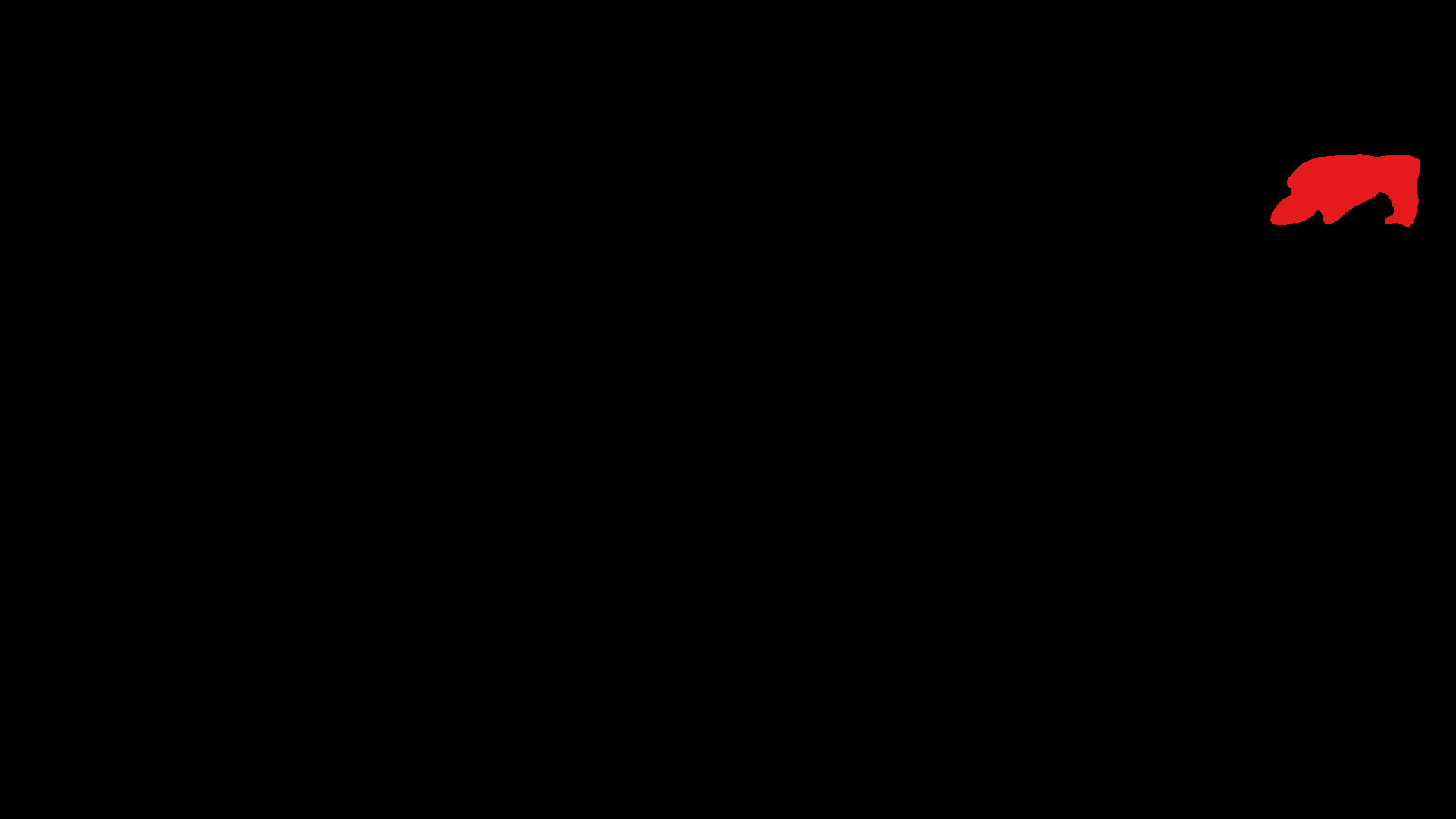}
        \caption{}
        \label{fig:sub-f}
    \end{subfigure}
    \vspace{0.1cm} 

    \begin{subfigure}[b]{0.3\linewidth}
        \centering
        \includegraphics[width=\linewidth]{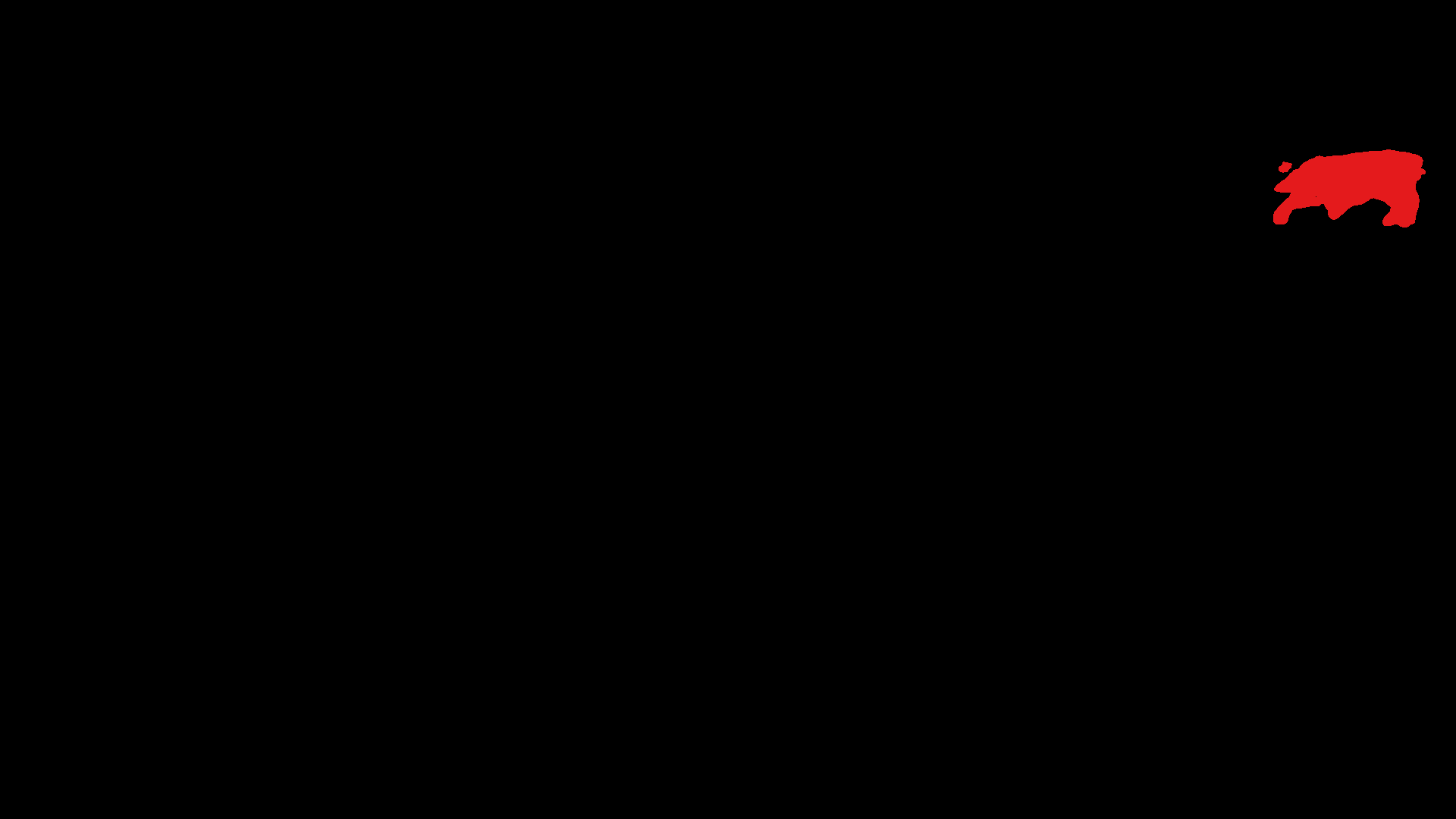}
        \caption{}
        \label{fig:sub-g}
    \end{subfigure}
    \hfill
    \begin{subfigure}[b]{0.3\linewidth}
        \centering
        \includegraphics[width=\linewidth]{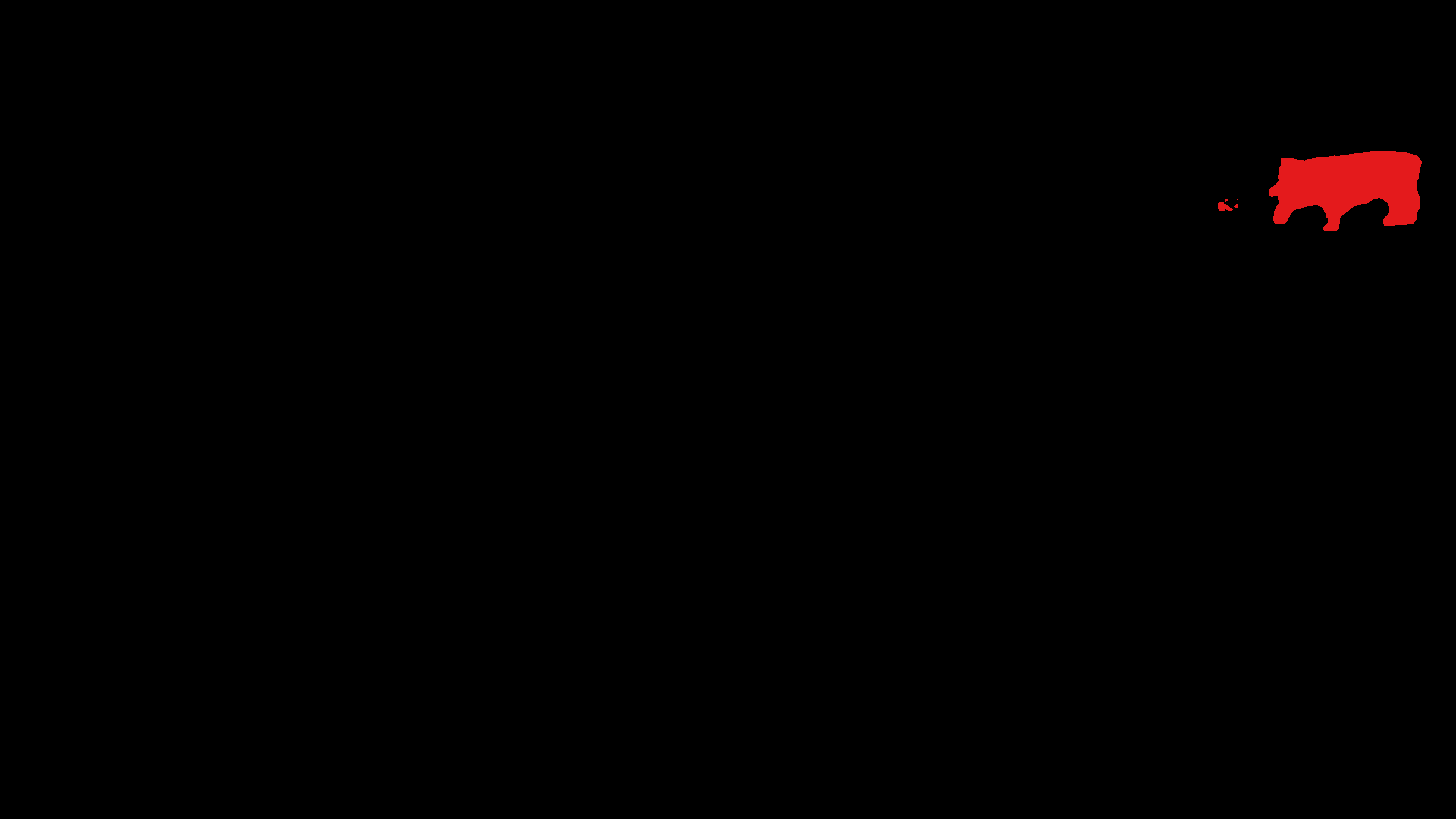}
        \caption{}
        \label{fig:sub-h}
    \end{subfigure}
    \hfill
    \begin{subfigure}[b]{0.3\linewidth}
        \centering
        \includegraphics[width=\linewidth]{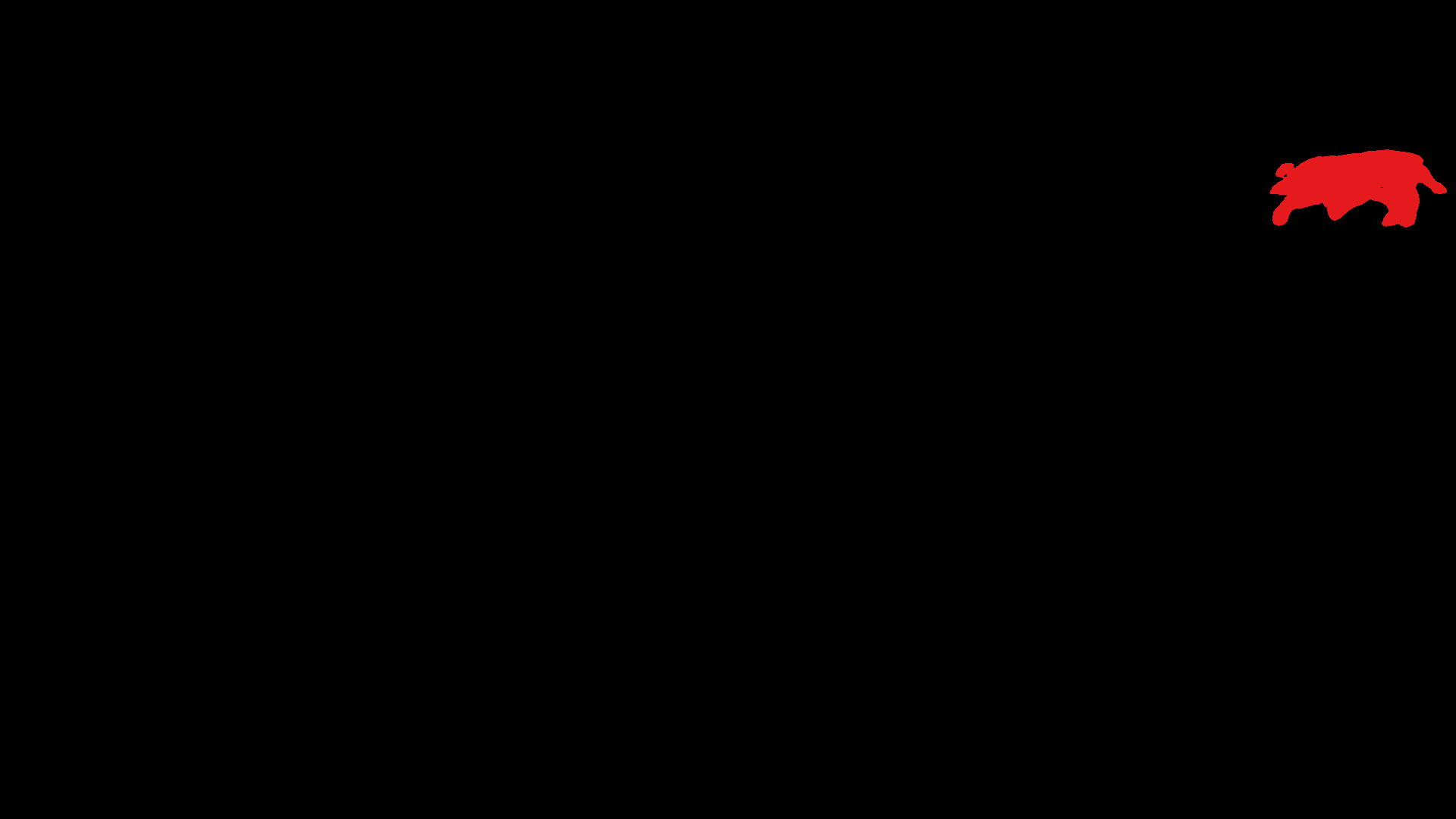}
        \caption{}
        \label{fig:sub-i}
    \end{subfigure}

    \caption{The inference outcomes of diverse models, along with our ultimate pseudo - label mask maps, are presented as follows. (a) represents the original RGB image. (b) is inferred by the TMO. (c) is the mask derived from the LiVOS .(d) is the mask obtained through the XMem. (e) is the mask inferred by the Cutie. (f) is the mask resulting from the inference of the SAM2. (g) is the average fusion mask map generated by converting the masks of the five models into bounding boxes and then performing an average fusion operation. (h) is the max fusion mask map. (i) is our pseudo-label. }
    \label{fig:gt} 
\end{figure}

The fused pseudo-label, as a key intermediate link, bridges the gap between the outputs of individual models and the performance of the unified optimization system.Fig \ref{fig:gt} presents a detailed fusion process along with various outcomes. It enables the intelligent selection of the model that demonstrates the best performance for different video contents. 

\subsubsection*{2.2.3 Model Recommendation Mechanism: Intelligent Task Allocation}
Based on the generated pseudo-label, we have developed a dynamic model recommendation mechanism to ensure that each video frame is processed by the most suitable model. 
\begin{itemize}
    \item Initially, feature extraction is conducted to analyze the video frames and extract key information, including scene complexity, the number of objects, and the distribution of object sizes.
    \item Subsequently, we have established a compact model performance database to record the historical performance of each model across various feature scenarios.
    \item Finally, a recommendation algorithm is employed to recommend the optimal model for each video frame based on the extracted frame features and the information stored in the model performance database.
\end{itemize}

By implementing this model recommendation mechanism, the system is able to dynamically allocate tasks to the most suitable model for each video frame, thereby enhancing the efficiency and accuracy of video frame segmentation and tracking tasks.

\section*{3. Experiments}

\subsection*{3.1 Dataset Introduction}
We fine-tune the segmentation model using the MOSE dataset. MOSE contains 2,149 video clips and 5,200 objects from 36 categories, with 431,725 high-quality object segmentation masks. The most significant feature of the MOSE dataset is its complex scenes with crowded or occluded objects. Target objects in videos are often occluded by other objects and disappear in certain frames.

\begin{figure*}[tb]
    \centering
    \begin{subfigure}[t]{0.9\textwidth}
        \centering
        \begin{adjustbox}{max width=\textwidth}
            \includegraphics[width=0.24\textwidth]{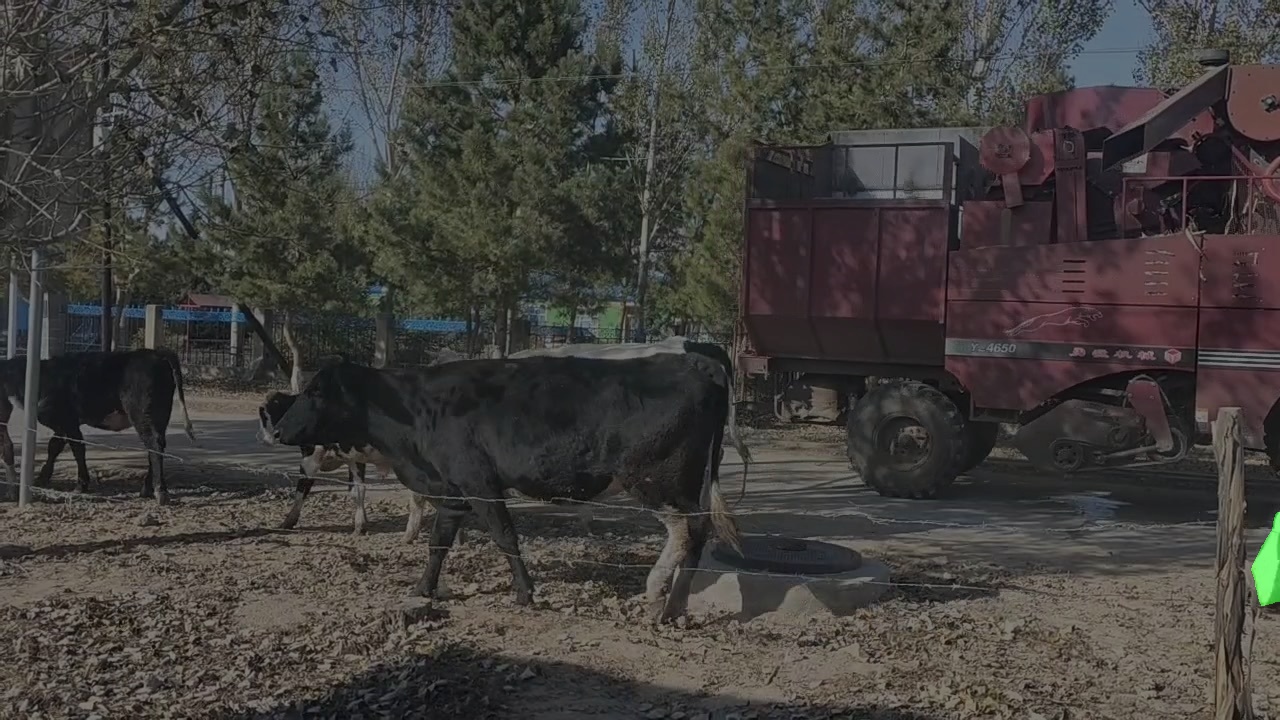}
            \includegraphics[width=0.24\textwidth]{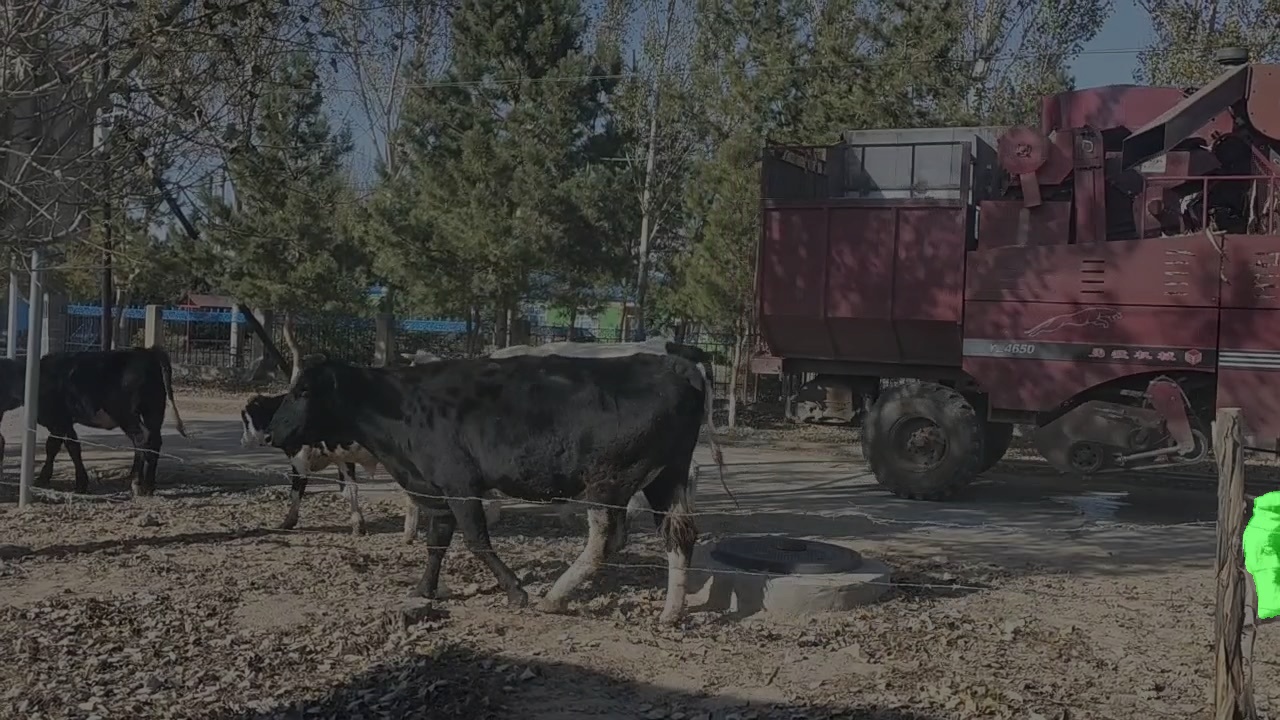}
            \includegraphics[width=0.24\textwidth]{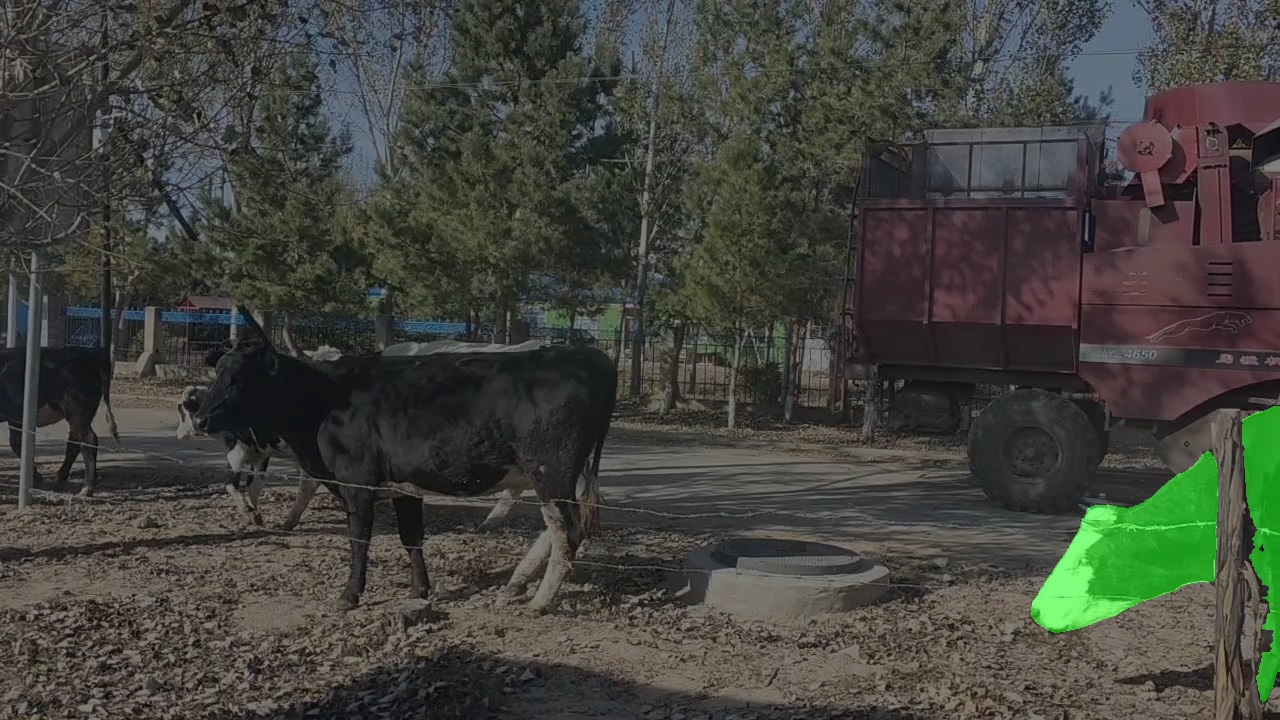}
            \includegraphics[width=0.24\textwidth]{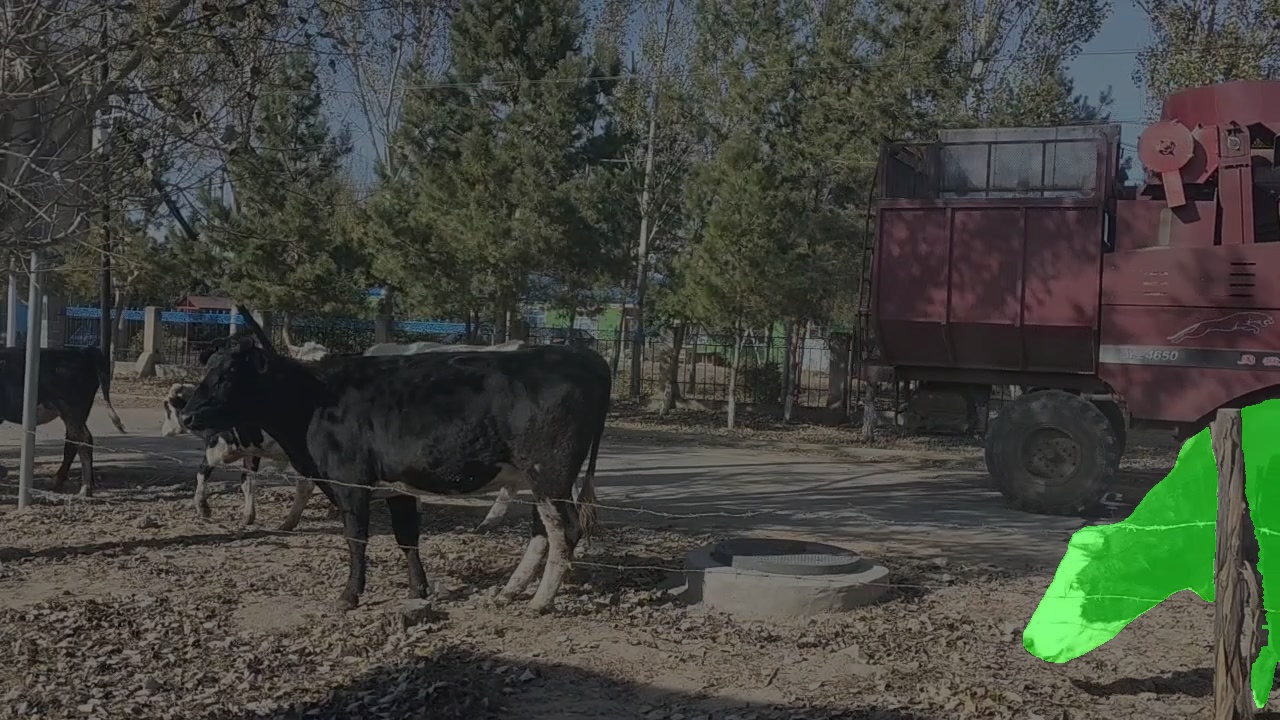}
        \end{adjustbox}
        \caption{}
    \end{subfigure}
    \vspace{0.3cm} 

    \begin{subfigure}[t]{0.9\textwidth}
        \centering
        \begin{adjustbox}{max width=\textwidth}
            \includegraphics[width=0.24\textwidth]{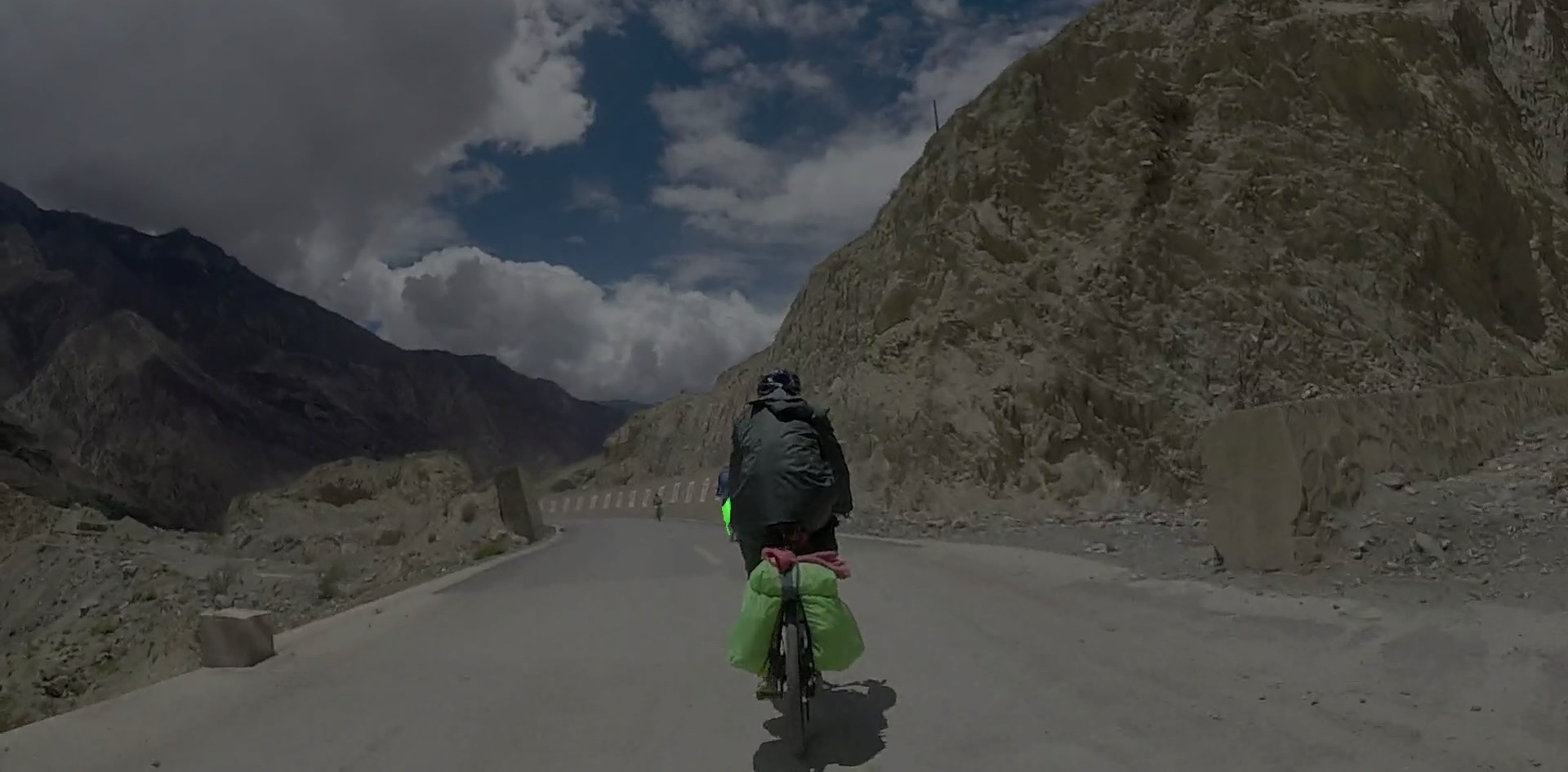}
            \includegraphics[width=0.24\textwidth]{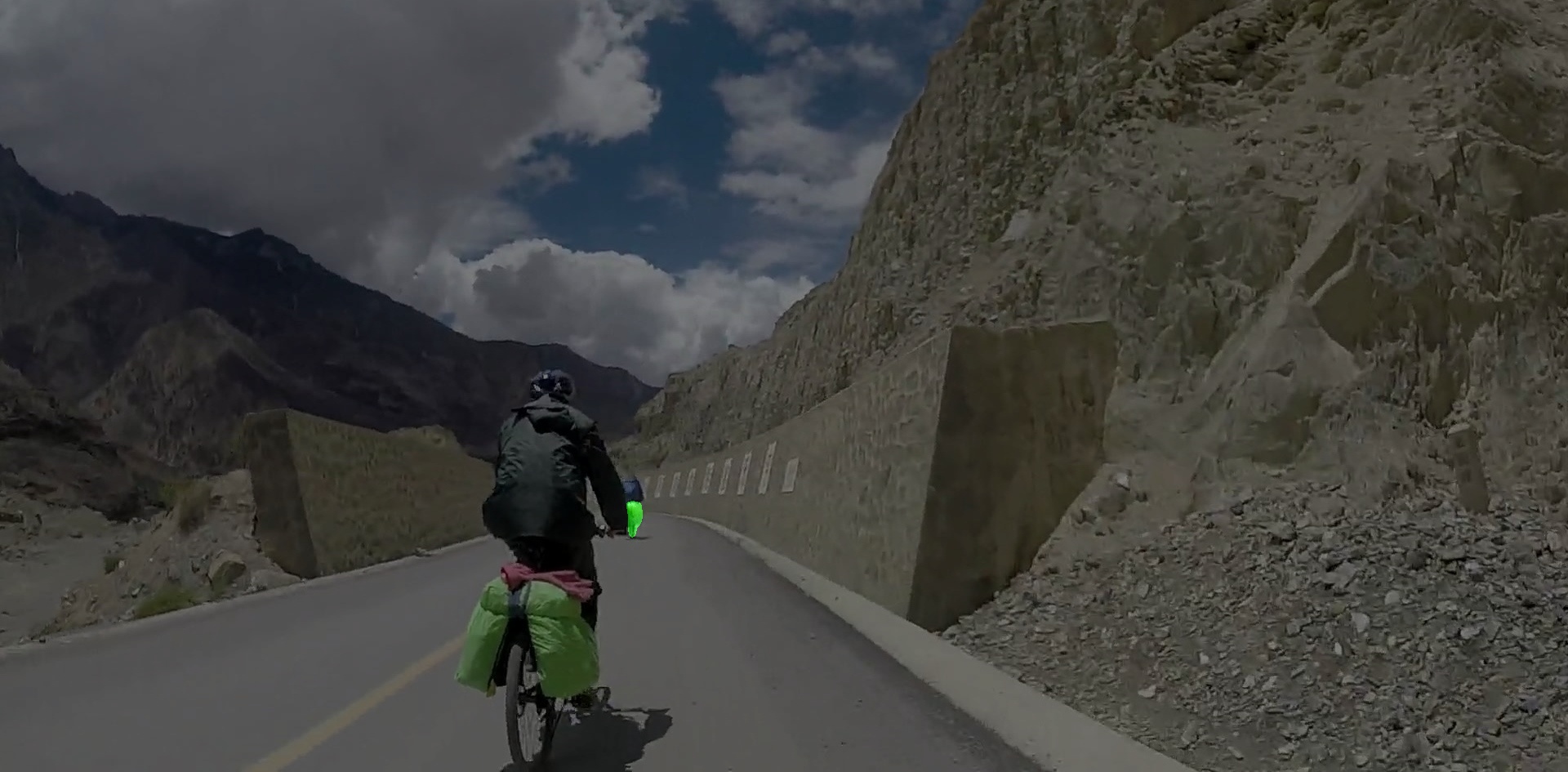}
            \includegraphics[width=0.24\textwidth]{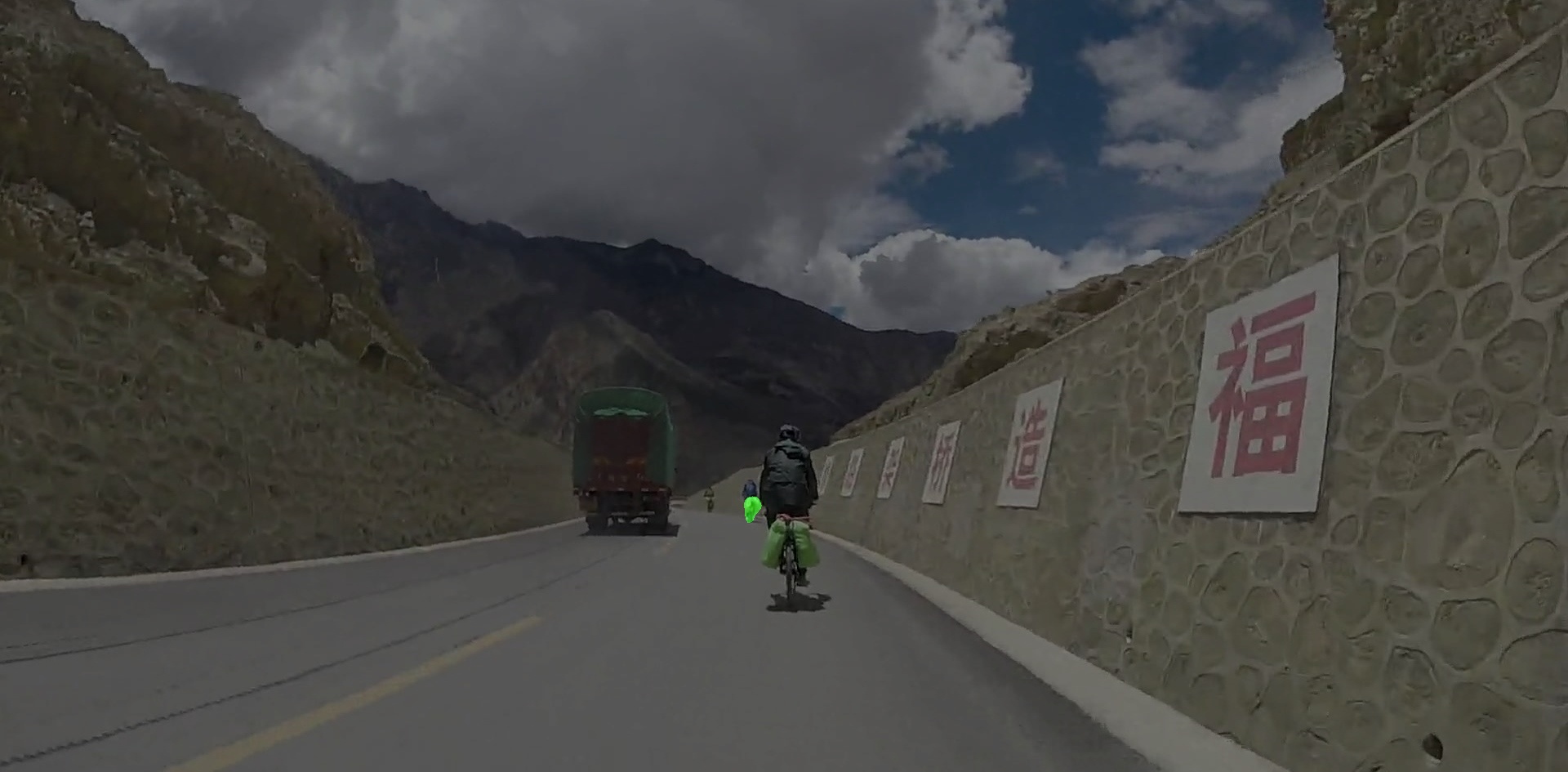}
            \includegraphics[width=0.24\textwidth]{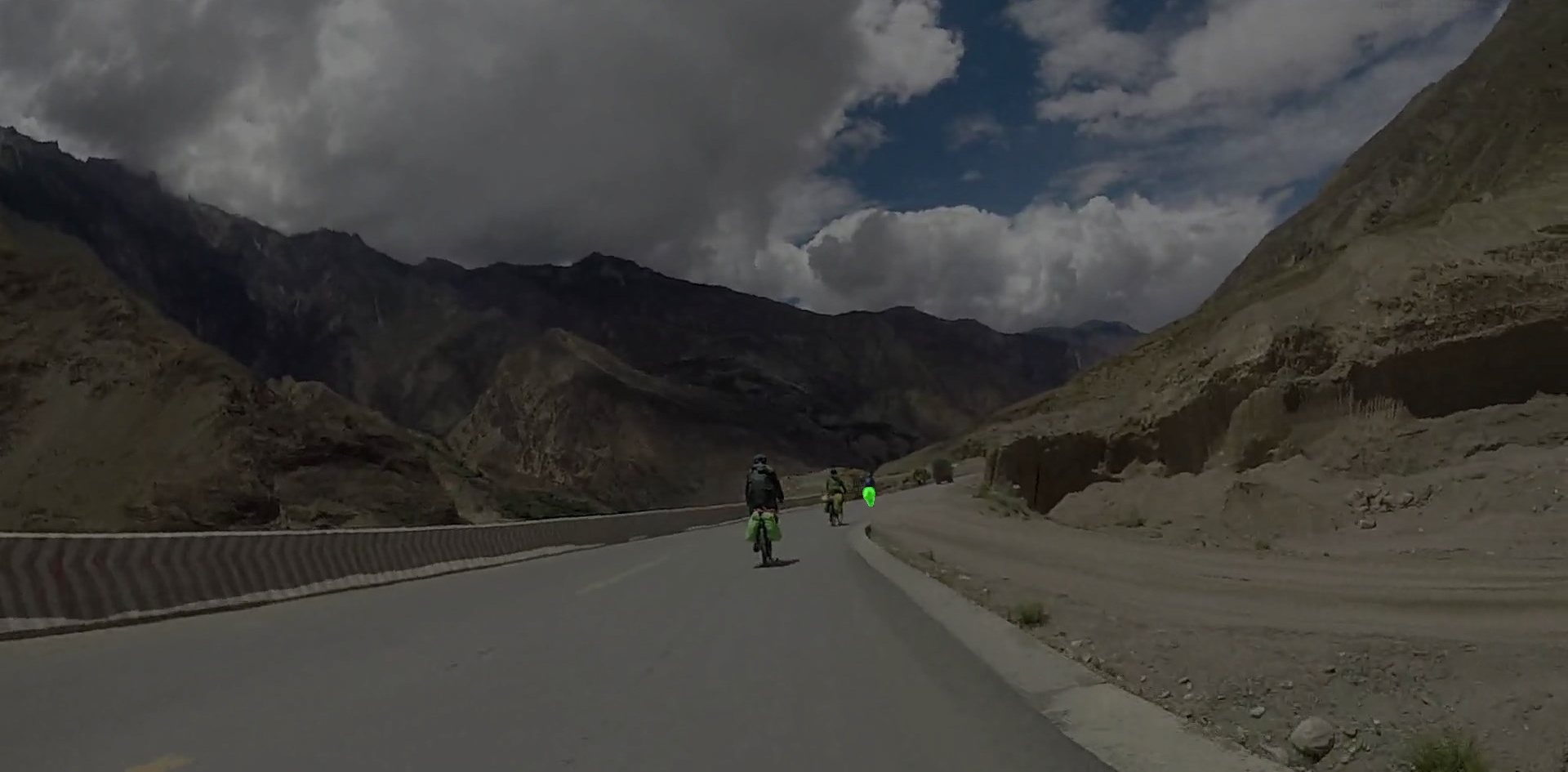}
        \end{adjustbox}
        \caption{}
    \end{subfigure}
    \vspace{0.3cm} 

    \begin{subfigure}[t]{0.9\textwidth}
        \centering
        \begin{adjustbox}{max width=\textwidth}
            \includegraphics[width=0.24\textwidth]{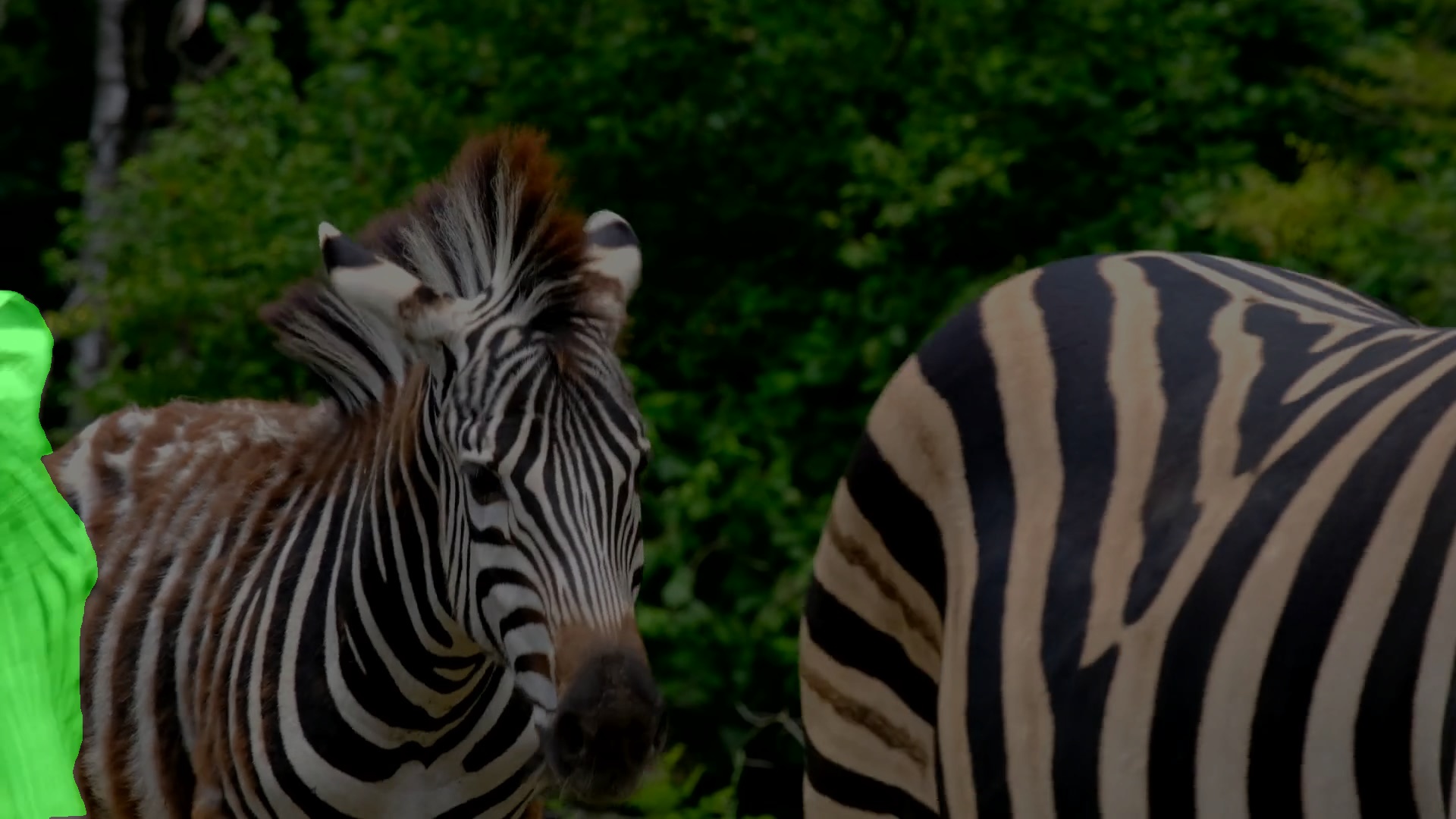}
            \includegraphics[width=0.24\textwidth]{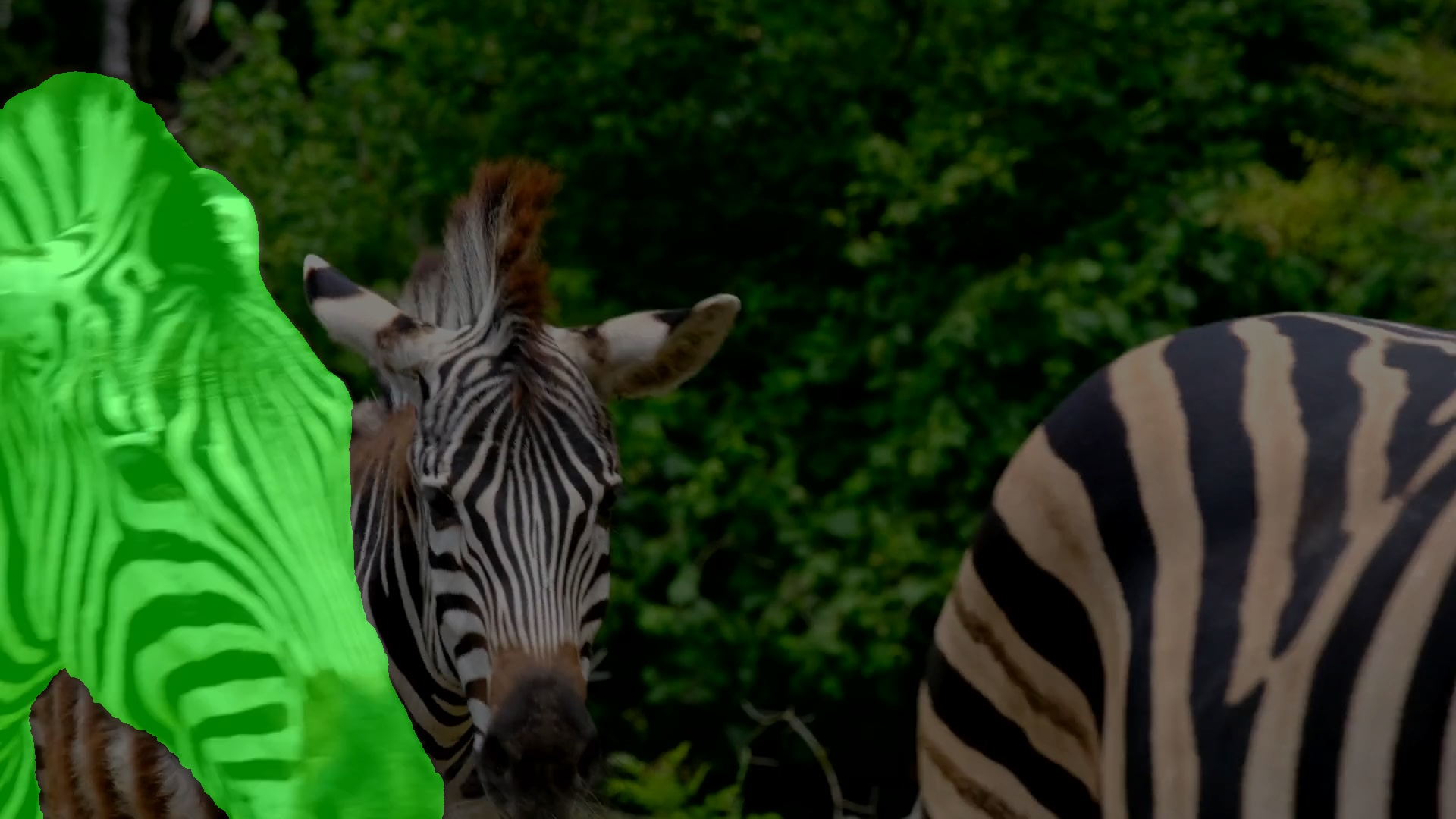}
            \includegraphics[width=0.24\textwidth]{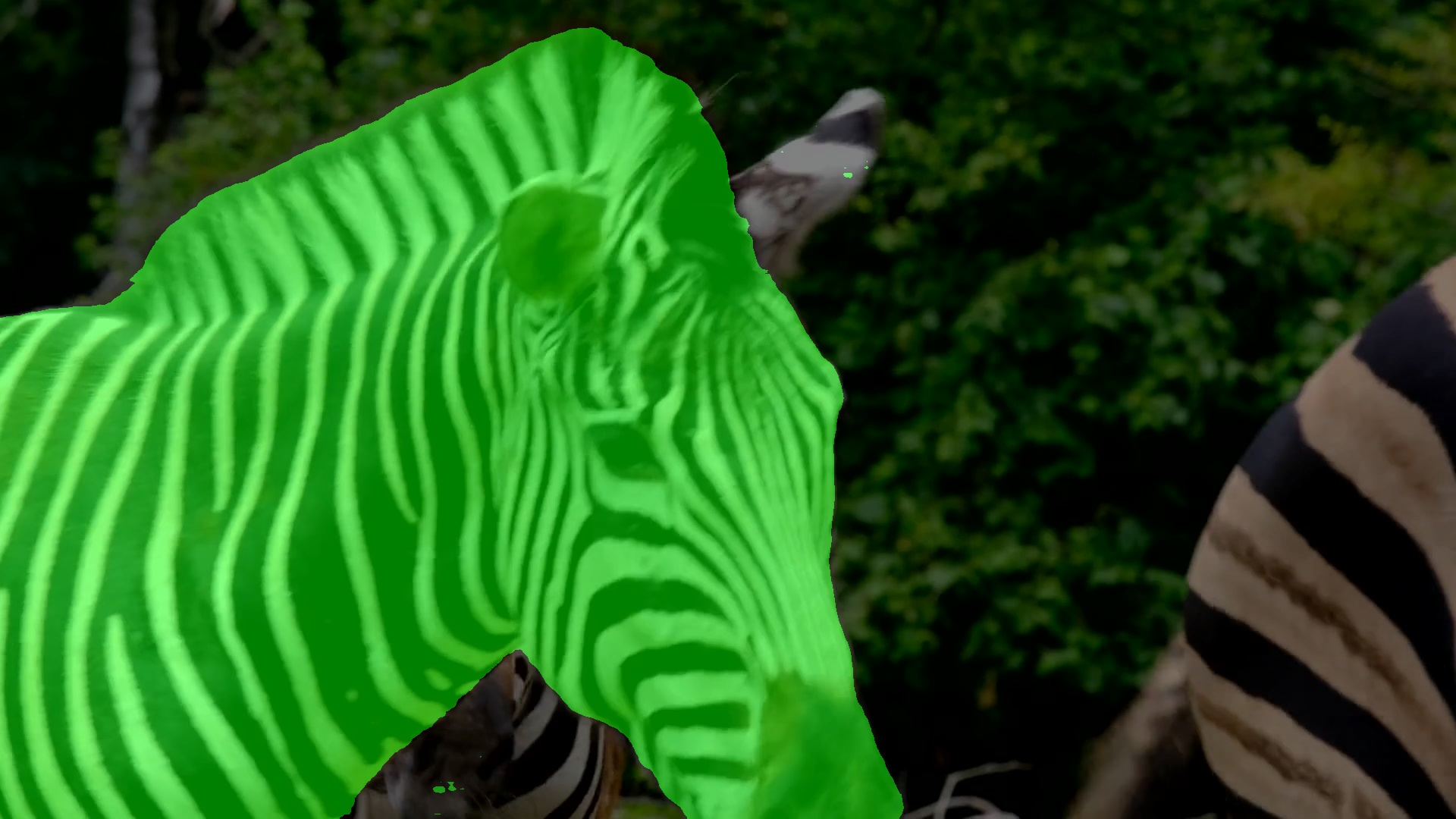}
            \includegraphics[width=0.24\textwidth]{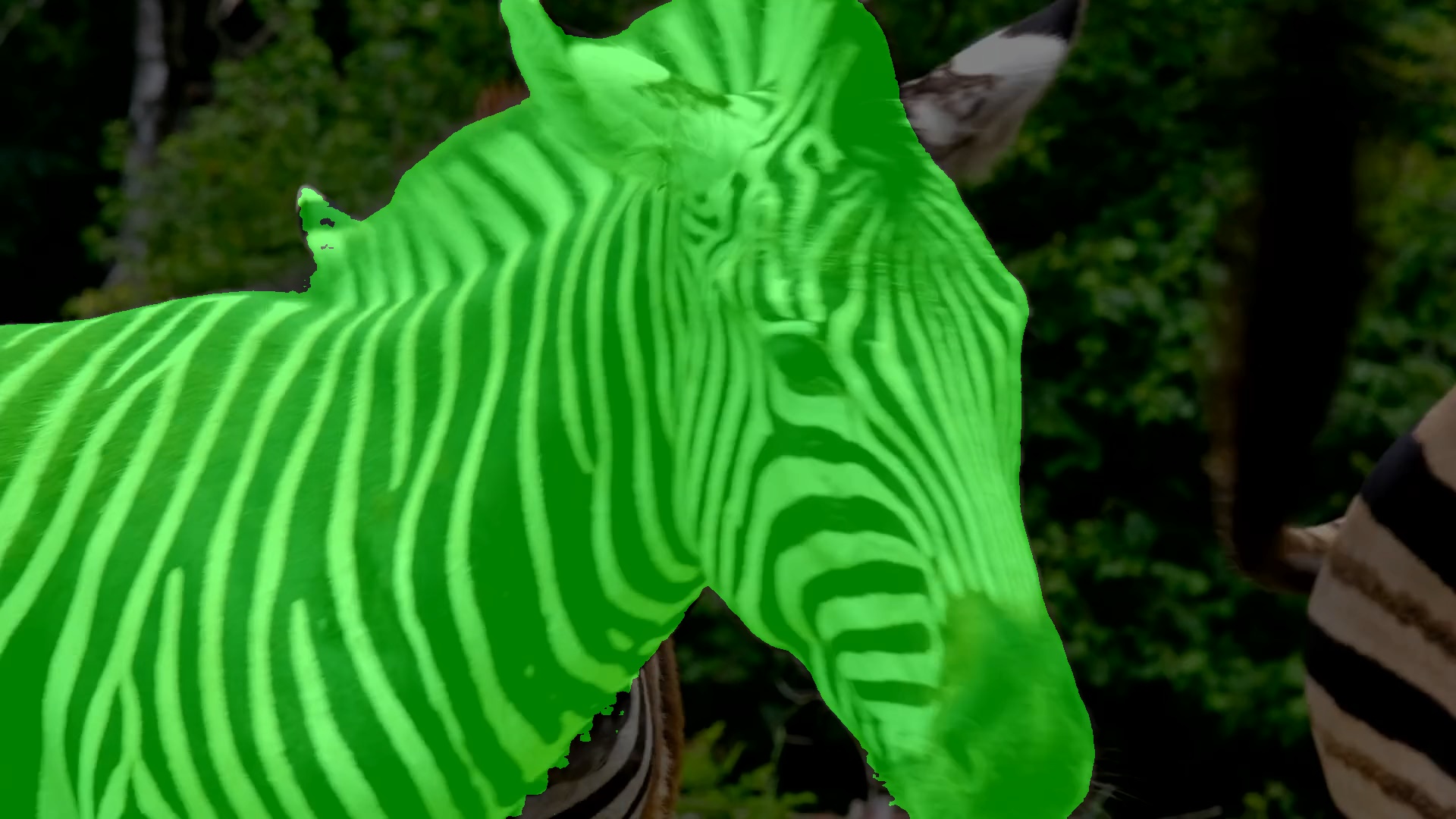}
        \end{adjustbox}
        \caption{}
    \end{subfigure}
    \vspace{0.3cm} 

    \begin{subfigure}[t]{0.9\textwidth}
        \centering
        \begin{adjustbox}{max width=\textwidth}
            \includegraphics[width=0.24\textwidth]{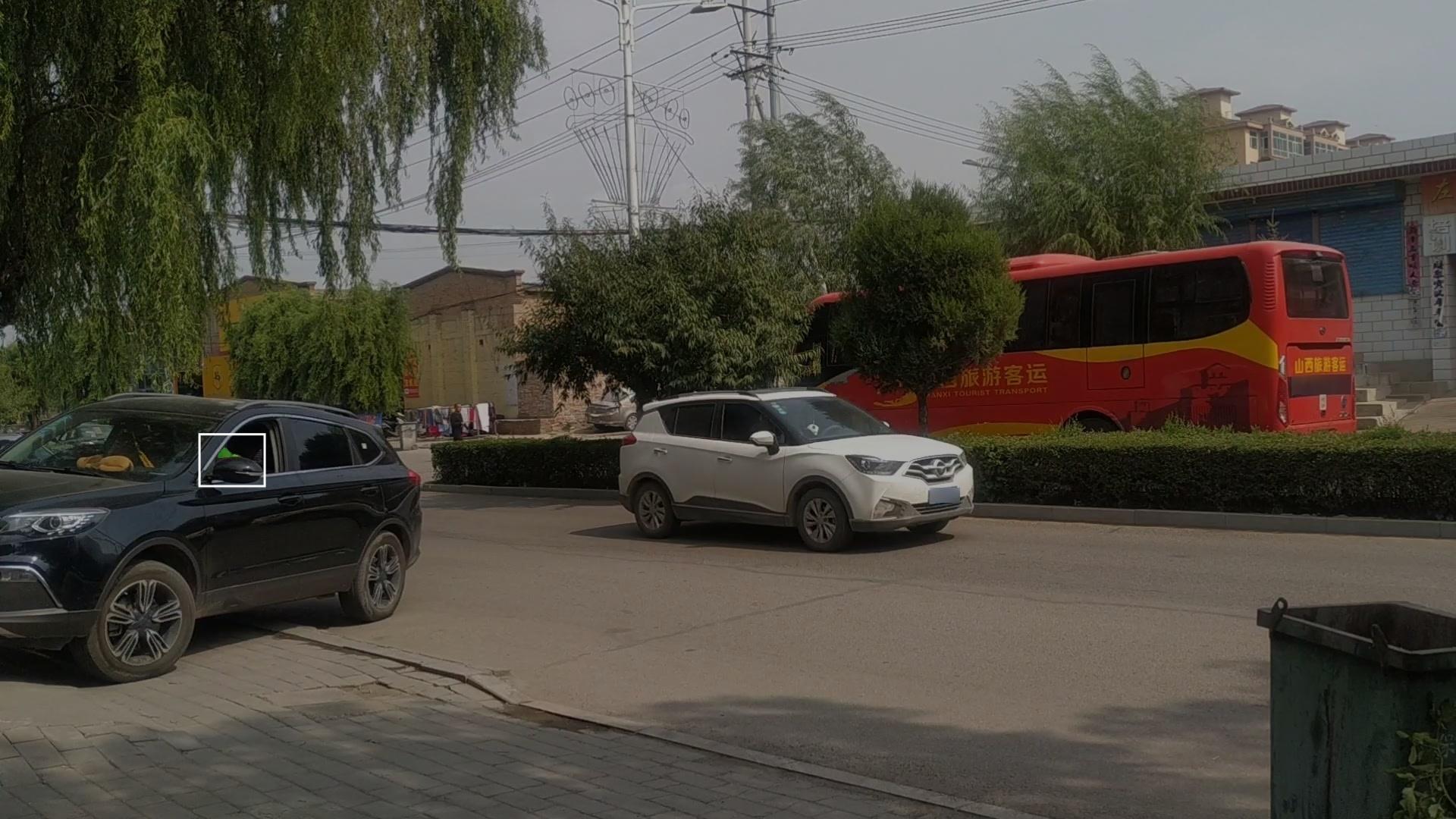}
            \includegraphics[width=0.24\textwidth]{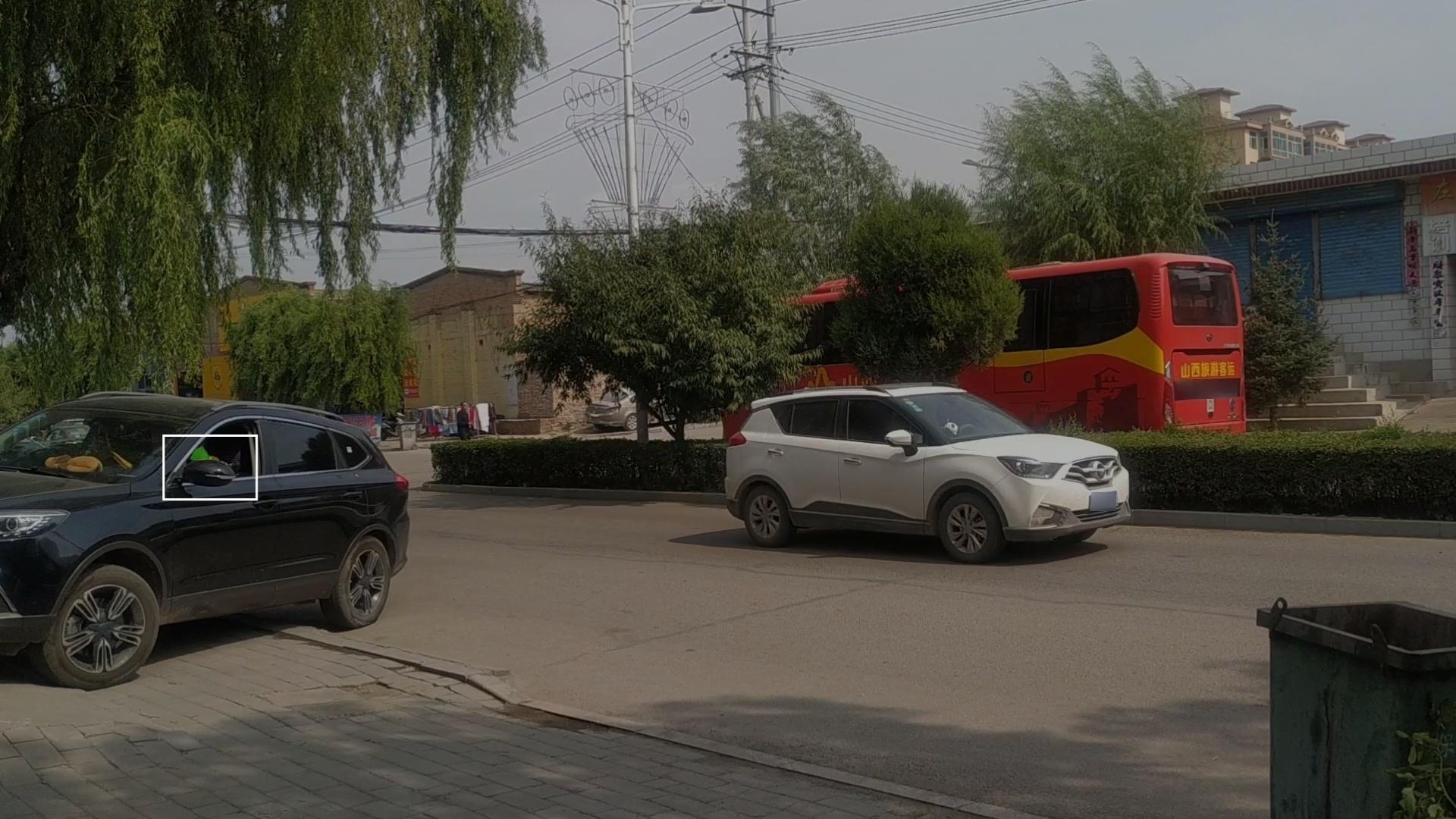}
            \includegraphics[width=0.24\textwidth]{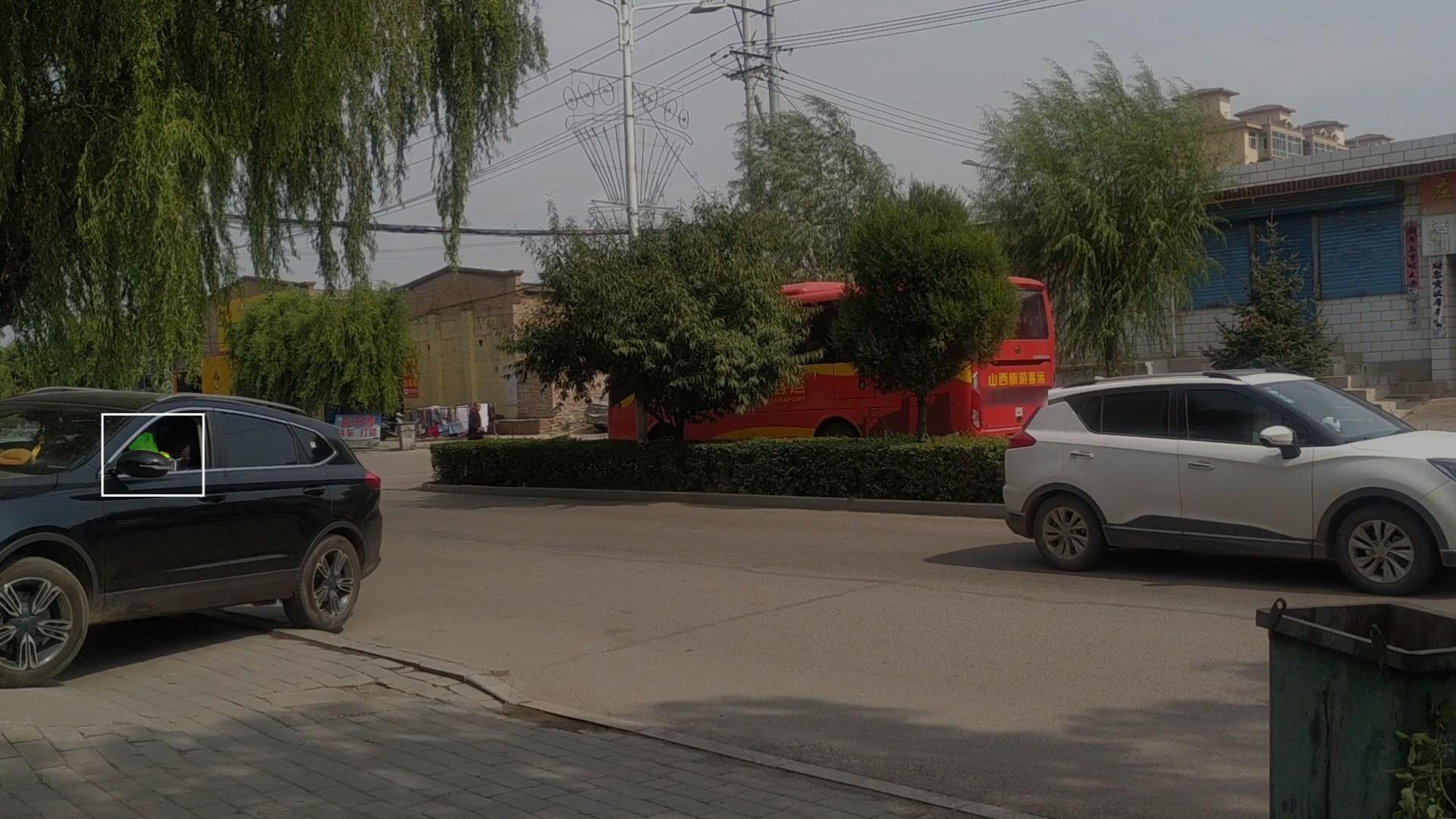}
            \includegraphics[width=0.24\textwidth]{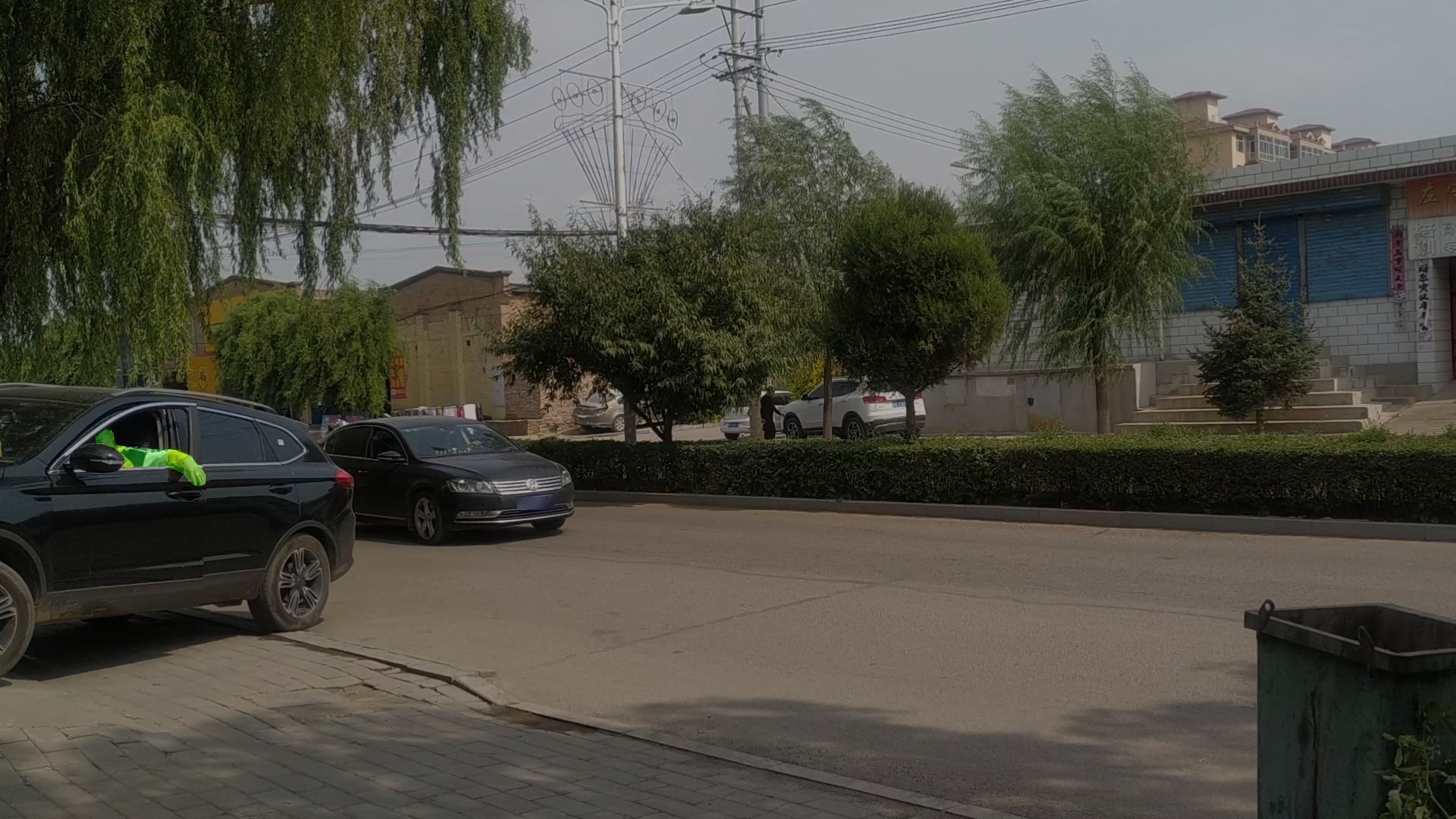}
        \end{adjustbox}
        \caption{}
    \end{subfigure}
    \caption{(a) demonstrates the outstanding robustness of the solution, showcasing its stable tracking ability for the target under extreme observation conditions.(b) reflects the powerful effectiveness of this proposed solution in dealing with severe occlusion situations.(c) emphasizes that our solution can clearly distinguish between similar objects, and the generated masks are extremely accurate.(d) exhibits that the solution can accurately segment the driver from the entire vehicle.}
    \label{fig:image} 
\end{figure*}

\subsection*{3.2 Evaluation Metrics}
We evaluate segmentation accuracy using the average Jaccard (J) index, average boundary F-score, and the average J\&F value. After inference by the recommended model, we assess its performance using the J\&F score (Joint and Fusion metric). The specific calculation methods are as follows:

\textbf{Segmentation Accuracy (J)} Compare the inferred results with the Ground Truth segmentation masks to calculate the IoU value, and take the average of all targets as the J score. For a predicted segmentation mask \( P \) and a ground truth segmentation mask \( G \), the Jaccard value is defined as:

\begin{equation}
J = \frac{|P \cap G|}{|P \cup G|} = \frac{\sum_i P_i \cdot G_i}{\sum_i P_i + \sum_i G_i - \sum_i P_i \cdot G_i},
\end{equation}

where \( P_i \) and \( G_i \) denote the value of the \( i \)-th pixel in the predicted and ground truth masks, respectively. The Jaccard value ranges from 0 to 1, with higher values indicating better performance.

\textbf{Tracking Effectiveness (F)}Compare the inferred results with the Ground Truth tracking IDs to calculate the MOTA (Multi-Object Tracking Accuracy) value, which measures the accuracy and stability of tracking.It is calculated as follows:

\begin{equation}
F = \frac{2 \cdot \text{Precision} \cdot \text{Recall}}{\text{Precision} + \text{Recall}},
\end{equation}

where 
\begin{equation}
\text{Precision} = \frac{|P \cap G|}{|P|} = \frac{\sum_i P_i \cdot G_i}{\sum_i P_i},
\end{equation}
and
\begin{equation}
\text{Recall} = \frac{|P \cap G|}{|G|} = \frac{\sum_i P_i \cdot G_i}{\sum_i G_i}.
\end{equation}

The F-Measure also ranges from 0 to 1, with higher values indicating better model performance in handling positive and negative samples.

\textbf{Composite Score} Weight J and F (e.g., 0.5:0.5) to derive the final J\&F score. A higher J\&F score indicates better overall model performance on the video frame.

\subsection*{3.3 Ablation Experiments}
To verify the performance of our solution, in the two finetuned models, we have all divided the same 15\% of the MOSE training set as the validation set.
\subsubsection*{3.3.1 Effect of Finetuning}
The experimental results on the validation set demonstrate that finetuning can improve the accuracy of the model.Taking SAM2 as an example, the comparison results before and after finetuning are presented in the Table. \ref{tab:fine_tuning}

\begin{table}[htbp]
    \centering
    \caption{Comparison before and after finetuning SAM2}
    \label{tab:fine_tuning}
    \begin{tabular}{cccc}
        \toprule
        Model & J & F & J\&F \\
        \midrule
        SAM2 & 78.63\% & 86.03\%  &  82.33\% \\
        SAM2-Fine\_tuning & 79.66\%  & 87.62\%  & 83.64\% \\
        \bottomrule
    \end{tabular}
\end{table}

\subsubsection*{3.3.2 Effect of Our Pseudo-Label Generation Strategy}
The experimental results in Table \ref{tab:PGMR} indicate that the Adaptive Pseudo-labels Guided Model Refinement Pipeline can significantly increase the J\&F score and effectively integrate the advantages of different models. averge method is the average fusion mask map generated by converting the masks of the five models into bounding boxes and then performing an average fusion operation. max method is the max fusion mask map obtained by converting the masks of the five models into bounding boxes and conducting a maximum fusion operation

\begin{table}[htbp]
    \centering
    \caption{Adaptive Pseudo-labels Guided Model Refinement Pipeline results}
    \label{tab:PGMR}
    \begin{tabular}{cccc}
        \toprule
        method & J & F & J\&F \\
        \midrule
        average & 82.76\%  & 89.20\%  & 85.98\% \\
        max & 80.57\%  & 88.09\%  & 84.33\% \\
        PGMR(ours) & 83.56\%  & 91.84\%  & 87.70\% \\
        \bottomrule
    \end{tabular}
\end{table}

\subsection*{3.4 Results}
Ultimately, our solution effectively leverages the strengths of different models to significantly enhance overall task performance. As shown in Table \ref{tab:leaderboard}, our proposed solution achieved 1st place in the complex video object segmentation track of the 2025 PVUW Challenge. Additionally, we present some quantitative results in Fig \ref{fig:image}. It can be seen that the proposed solution can accurately segment small objects and distinguish similar targets.
\begin{table}[htbp]
    \centering
    \caption{Leaderboard during the test phase}
    \label{tab:leaderboard}
    \begin{tabular}{lccc} 
        \toprule 
        \textbf{User} & \textbf{J} & \textbf{F} & \textbf{J\&F} \\ 
        \midrule
        imaplus & 0.8359 & 0.9092 & 0.8726 \\ 
        KirinCZW & 0.8250 & 0.9007 & 0.8628 \\ 
        dumplings & 0.8028 & 0.8757 & 0.8392 \\ 
        SCU\_Leung & 0.7993 & 0.8733 & 0.8363 \\ 
        wulutulumuman & 0.7989 & 0.8721 & 0.8355 \\ 
        \bottomrule
    \end{tabular}
\end{table}

\section*{4. Conclusion}
STSeg integrates the advantageous modules of SAM and TMO, is finetuned on the MOSE dataset, and applies data augmentation and Adaptive Pseudo-labels Guided Model Refinement Pipeline  during inference. Ultimately, we achieved a J\&F score of 87.26\% and secured first place in MOSE Track of the 4th PVUW challenge.

\bibliographystyle{unsrt}  
\bibliography{references}  

\begin{thebibliography}{10}

\bibitem{ding2023mose}
Henghui Ding, Chang Liu, Shuting He, Xudong Jiang, Philip~HS Torr, and Song Bai.
\newblock Mose: A new dataset for video object segmentation in complex scenes.
\newblock In {\em Proceedings of the IEEE/CVF international conference on computer vision}, pages 20224--20234, 2023.

\bibitem{ravi2024sam}
Nikhila Ravi, Valentin Gabeur, Yuan-Ting Hu, Ronghang Hu, Chaitanya Ryali, Tengyu Ma, Haitham Khedr, Roman R{\"a}dle, Chloe Rolland, Laura Gustafson, et~al.
\newblock Sam 2: Segment anything in images and videos.
\newblock {\em arXiv preprint arXiv:2408.00714}, 2024.

\bibitem{cho2023treating}
Suhwan Cho, Minhyeok Lee, Seunghoon Lee, Chaewon Park, Donghyeong Kim, and Sangyoun Lee.
\newblock Treating motion as option to reduce motion dependency in unsupervised video object segmentation.
\newblock In {\em Proceedings of the IEEE/CVF winter conference on applications of computer vision}, pages 5140--5149, 2023.

\bibitem{Ding2024PVUW}
Henghui Ding, Chang Liu, Yunchao Wei, Nikhila Ravi, Shuting He, Song Bai, Philip H.~S. Torr, Deshui Miao, Xin Li, Zhenyu He, et~al.
\newblock Pvuw 2024 challenge on complex video understanding: Methods and results.
\newblock {\em arXiv preprint}, arXiv:2406.17005, 2024.

\bibitem{Ding_2023_ICCV}
Henghui Ding, Chang Liu, Shuting He, Xudong Jiang, and Chen~Change Loy.
\newblock Mevis: A large-scale benchmark for video segmentation with motion expressions.
\newblock In {\em Proceedings of the IEEE/CVF International Conference on Computer Vision (ICCV)}, pages 2694--2703, October 2023.

\bibitem{cheng2023putting}
Ho~Kei Cheng, Seoung~Wug Oh, Brian Price, Joon-Young Lee, and Alexander Schwing.
\newblock Putting the object back into video object segmentation.
\newblock {\em arXiv preprint arXiv}, 2023.
\newblock 2023a.

\bibitem{cheng2021modular}
Ho~Kei Cheng, Yu-Wing Tai, and Chi-Keung Tang.
\newblock Modular interactive video object segmentation: Interaction-to-mask, propagation and difference-aware fusion.
\newblock In {\em CVPR}, 2021.
\newblock 2021b.

\bibitem{seong2020kernelized}
Hongje Seong, Junhyuk Hyun, and Euntai Kim.
\newblock Kernelized memory network for video object segmentation.
\newblock In {\em European Conference on Computer Vision}, pages 629--645. Springer, 2020.

\bibitem{oh2019video}
Seoung~Wug Oh, Joon-Young Lee, Ning Xu, and Seon~Joo Kim.
\newblock Video object segmentation using space-time memory networks.
\newblock In {\em Proceedings of the IEEE/CVF International Conference on Computer Vision}, pages 9226--9235, 2019.

\bibitem{zhang2023joint}
Jiaming Zhang, Yutao Cui, Gangshan Wu, and Limin Wang.
\newblock Joint modeling of feature, correspondence, and a compressed memory for video object segmentation.
\newblock {\em arXiv preprint arXiv:2308.13505}, 2023.
\newblock 2023b.

\bibitem{cheng2021rethinking}
Ho~Kei Cheng, Yu~Wing Tai, and Chi~Keung Tang.
\newblock Rethinking space - time networks with improved memory coverage for efficient video object segmentation.
\newblock In {\em NeurIPS}, 2021.
\newblock 2021a.

\bibitem{cheng2022xmem}
Ho~Kei Cheng and Alexander~G Schwing.
\newblock Xmem: Long-term video object segmentation with an atkinson-shiffrin memory model.
\newblock In {\em European Conference on Computer Vision}, pages 640--658. Springer, 2022.

\bibitem{pont20172017}
Jordi Pont-Tuset, Federico Perazzi, Sergi Caelles, Pablo Arbel{\'a}ez, Alex Sorkine-Hornung, and Luc Van~Gool.
\newblock The 2017 davis challenge on video object segmentation.
\newblock {\em arXiv preprint arXiv:1704.00675}, 2017.

\bibitem{xu2018youtube}
Ning Xu, Linjie Yang, Yuchen Fan, Dingcheng Yue, Yuchen Liang, Jianchao Yang, and Thomas Huang.
\newblock Youtube-vos: A large-scale video object segmentation benchmark.
\newblock {\em arXiv preprint arXiv:1809.03327}, 2018.

\bibitem{videnovic2024distractor}
Jovana Videnovic, Alan Lukezic, and Matej Kristan.
\newblock A distractor-aware memory for visual object tracking with sam2.
\newblock {\em arXiv preprint arXiv:2411.17576v1}, 2024.
\newblock License: CC BY 4.0; Submitted on 26 Nov 2024.

\end{thebibliography}
\end{document}